\documentclass[twocolumn,10pt,cleanhead,cleanfoot]{asme2e}

\usepackage{amssymb}
\usepackage{graphicx}
\usepackage{caption}
\usepackage{subcaption}
\usepackage{amsmath}
\usepackage{xcolor}

\usepackage{algorithm}
\usepackage{algpseudocode}
\usepackage{hyperref}
\usepackage[utf8]{inputenc}
\usepackage{booktabs}
\usepackage{amsmath}
\usepackage{amssymb}
\usepackage{array}
\usepackage[most]{tcolorbox}
\definecolor{burntorange}{RGB}{191, 87, 0}
\definecolor{burntorangelight}{RGB}{255, 243, 230}
\newtcolorbox{llmbox}[1][]{%
    enhanced,
    breakable,
    frame hidden,
    colback=white,
    boxrule=0pt,
    sharp corners,
    left=10pt, right=4pt, top=4pt, bottom=4pt,
    borderline west={1.2pt}{0pt}{black},
    fontupper=\footnotesize\itshape,
    before upper={\setlength{\parskip}{4pt}\setlength{\parindent}{0pt}},
    coltitle=black,
    colbacktitle=white,
    fonttitle=\footnotesize\sffamily\bfseries\upshape,
    halign title=flush left,
    attach title to upper={\par\vspace{2pt}},
    #1
}

\usepackage[utf8]{inputenc}
\usepackage{booktabs}
\usepackage{amsmath}
\usepackage{amssymb}
\usepackage{array}
\usepackage[most]{tcolorbox}
\definecolor{burntorange}{RGB}{191, 87, 0}
\definecolor{burntorangelight}{RGB}{255, 243, 230}

\definecolor{softred}{RGB}{160, 50, 50}          
\definecolor{softredlight}{RGB}{252, 235, 235}   

\newtcolorbox{agentquote}[1]{%
    enhanced,
    breakable,
    frame hidden,
    colback=white,
    boxrule=0pt,
    sharp corners,
    left=10pt, right=4pt, top=2pt, bottom=2pt,
    borderline west={1.2pt}{0pt}{black},
    fontupper=\footnotesize\itshape,
    before upper={\setlength{\parskip}{4pt}\setlength{\parindent}{0pt}},
}

\definecolor{designerblue}{RGB}{40, 90, 150}        
\definecolor{designerbluelight}{RGB}{232, 240, 252} 
\definecolor{constraintred}{RGB}{180, 60, 60}       
\definecolor{loadgreen}{RGB}{40, 130, 70}           

\newtcolorbox{designerbox}[1][]{%
    enhanced,
    sharp corners,
    colback=designerbluelight,
    colframe=designerblue,
    boxrule=0.8pt,
    left=8pt, right=8pt, top=6pt, bottom=6pt,
    fontupper=\footnotesize,
    before upper={\setlength{\parskip}{4pt}\setlength{\parindent}{0pt}},
    #1
}

\newtcolorbox{llmboxred}[1][]{%
    enhanced,
    breakable,
    frame hidden,
    colback=white,
    boxrule=0pt,
    sharp corners,
    left=10pt, right=4pt, top=4pt, bottom=4pt,
    borderline west={1.2pt}{0pt}{softred},
    fontupper=\footnotesize\itshape,
    before upper={\setlength{\parskip}{4pt}\setlength{\parindent}{0pt}},
    coltitle=softred,
    colbacktitle=white,
    fonttitle=\footnotesize\sffamily\bfseries\upshape,
    halign title=flush left,
    attach title to upper={\par\vspace{2pt}},
    #1
}



\title{TO-Agents: A Multi-Agent AI Framework for Subjective Preference-Guided Topology Optimization}
\usepackage{helvet}

\author{Isabella A. Stewart$^{1}$ \quad Hongrui Chen$^{2}$ \quad Faez Ahmed$^{2}$\thanks{Address all correspondences to istewart@mit.edu}
    \affiliation{
        $^1$Department of Civil and Environmental Engineering\\
        Massachusetts Institute of Technology\\
        Cambridge, MA, 02139 USA}
    \affiliation{
        $^2$Department of Mechanical Engineering\\
        Massachusetts Institute of Technology\\
        Cambridge, MA, 02139 USA}
}

\begin{document}

\maketitle    

\begin{abstract}

\textit{Topology optimization can generate efficient structures, but designers often must manually translate qualitative intent, such as desired visual style, product experience, or manufacturability into solver settings that are not directly tied to those preferences. We present TO-Agents, a multi-agent AI framework that connects natural-language design intent with iterative topology optimization. The framework converts a human-provided problem description into validated solver inputs, runs a topology optimization solver, renders the resulting 3D topology, and uses multiview vision-language reasoning with an independent judge agent to critique each result and revise solver parameters. We evaluate the framework on two long-horizon design tasks: a cantilever beam benchmark and a phone-stand product design. In both tasks, the designer specifies an aesthetic preference for hierarchically branched structures inspired by natural tree morphologies, and the system performs four revision cycles across ten independent replicates. TO-Agents produces at least one preference-aligned design in 60\% of trials for each case study, corresponding to up to $6 \times$ more successful trials than an ablated pipeline without visual or historical feedback. Judge scores and human evaluations show that the pipeline can identify effective parameter levers, recover from poor revisions, and expand design exploration. A manufacturing agent further post-processes top-ranked designs for additive manufacturing, enabling end-to-end intent-to-prototype design. We also identify failure modes, including overshooting, selective memory, misplaced tools, and incorrect parameter reasoning. These results suggest that agentic topology optimization can shift designers from low-level parameter tuning toward higher-level specification of form and function, while highlighting safeguards needed for reliable autonomous engineering design. Code is available at \url{https://github.com/bellastewart/TO-Agents}}. 

\end{abstract}


\section{INTRODUCTION}

Engineering design is fundamentally shaped by human preferences concerning performance, cost, manufacturability, appearance, and the tradeoffs among them. Translating these preferences into a viable product, however, requires substantial human effort. Because preferences are often incomplete, implicit, or contested, designers, engineers, and manufacturers must repeatedly negotiate desired outcomes and revise the means of achieving them. For engineers, this process often entails coordinating parameter tuning across siloed tools and workflows \cite{schiffer_hitop_2023, mueller_combining_2015}, while managing long-horizon processes and competing objectives \cite{krish_practical_2011, lazarov_length_2016, regenwetter_multi-objective_2025}.

\subsection{When Topology Optimization Meets Human Preferences}

Computational design methods reduce some of this effort by automating the exploration of large design spaces. Topology optimization (TO), for example, determines how material should be distributed within a design domain to optimize performance, such as maximizing stiffness subject to a prescribed volume fraction. In doing so, it can uncover high-performing and non-intuitive designs that may be inaccessible through engineering intuition alone.

Yet automation does not eliminate the need to translate human preferences into computational terms. Engineers must select objectives such as compliance, mass, or thermal response. They must also define loads, boundary conditions, material behavior, geometric limits, and manufacturability constraints \cite{sigmund_99_2001, andreassen_efficient_2011}, as well as configure parameters specific to the optimization algorithm \cite{sigmund_benchmarking_2022}. Because the resulting designs are highly sensitive to these choices, engineers often need to reformulate and rerun the optimization repeatedly to determine which outcomes best reflect their intent.

The difficulty is greatest when preferences are semantic, subjective, or context-dependent and therefore cannot be expressed adequately as fixed objectives or constraints. Aesthetic preferences, for instance, often concern qualities that emerge from interactions among multiple design decisions rather than from any single controllable parameter. Engineers must consequently evaluate generated designs, infer how parameter changes might produce more desirable outcomes, and repeat the optimization. They may otherwise need to rely on ad hoc post-processing and compromise \cite{oduguwa_handling_2003, brintrup_interactive_2007}. Human preference evaluation therefore becomes a central bottleneck that limits the scalability of computational design, introduces fatigue and subjective inconsistency, and reduces throughput in workflows that depend on extensive iterative exploration.

\subsection{The Growing Need for Human-Centered Topology Optimization}
We see growing importance in these considerations as topology optimization expands beyond purely performance-driven products into human-centered product design. In these applications, aesthetics, personalization, and other human considerations are integral to the design outcome and become the central pain-point of this long-horizon design process. Examples span high-performance equipment, patient-specific medical devices, and consumer products that balance engineering, ergonomic, and aesthetic requirements. 


The prevalence of such products are growing alongside the democratization of affordable 3D printing, which has made topology optimization increasingly accessible to hobbyists and everyday users. Platforms such as Thingiverse already feature simple topology-optimized everyday products designed by non-experts, including phone stands, laptop stands, bookends, drone frames, and shelf brackets. Such social platforms allow non-experts to experiment with these designs and share feedback reflecting their frustrations and preferences; however, the volume of comments and the technical complexity of TO make manually incorporating this feedback impractical. 

Consequently, an emerging area of topology optimization research aims to bridge its physics-driven nature with human preferences, making it easier for people to incorporate topology optimization into their designs, as discussed in the following related-work section.

\textbf{Yet, topology optimization still lacks a direct way to incorporate semantic design information.} Making it easier for designers and TO-novices to translate their ideas into optimization problems could reduce expertise barriers and further democratize topology optimization. 


\begin{figure*}[t]
    \centering
    \includegraphics[width=1\linewidth]{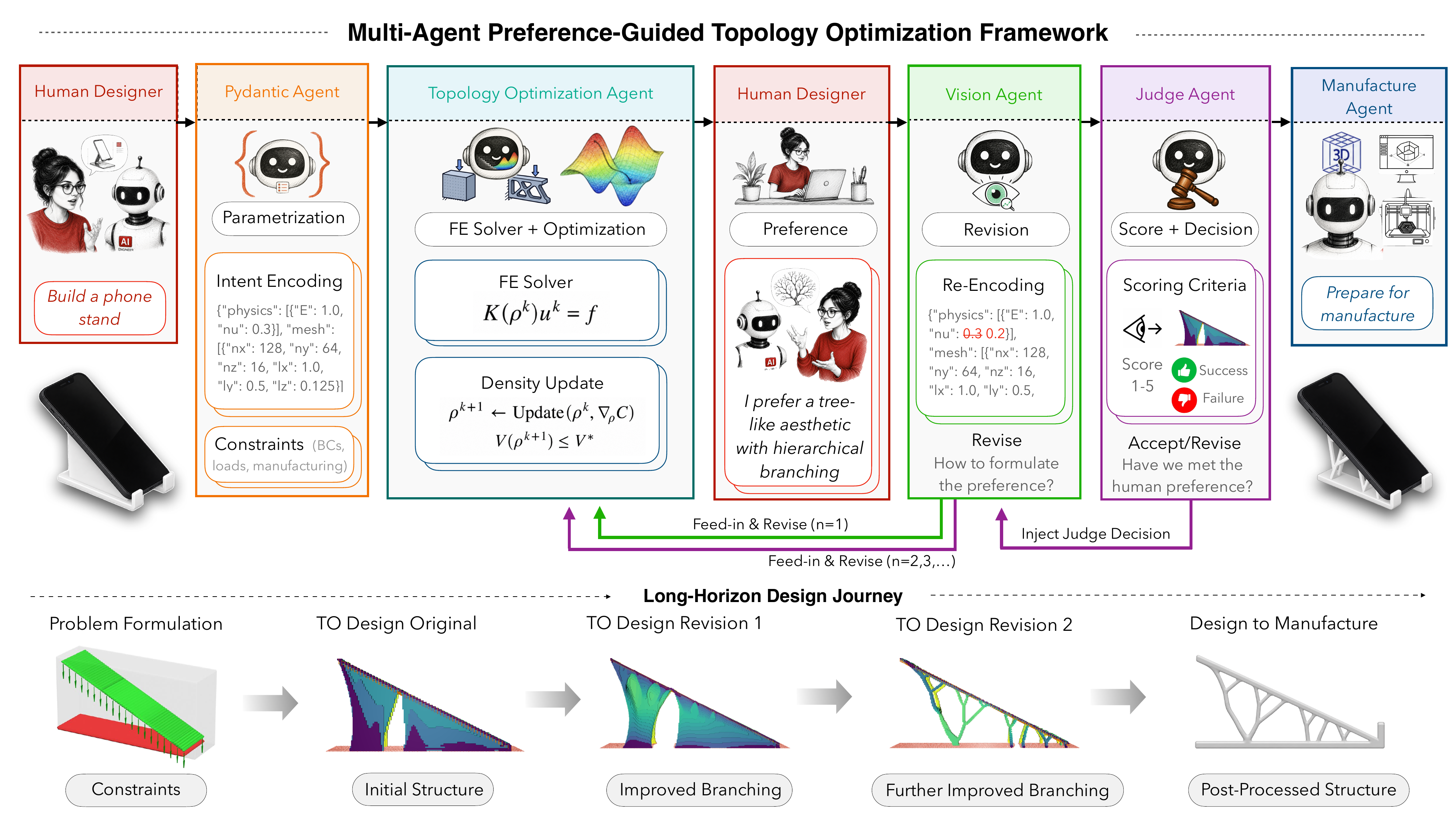}
    \caption{The TO-Agents pipeline consisting of a sequence of subagents that interact with a human designer in a multi-turn conversation architecture. Agents iteratively adjust the design parameters until the resulting structure satisfies the human designer’s qualitative objective of a bioinspired, dendriform architecture with hierarchical branching. Injection of judge agent feedback enables the system toward this design goal as the number of revisions grows.}
    \label{fig:pipeline}
\end{figure*}

\subsection{Translating Human Preferences into Optimization with LLMs and Agentic AI}

We posit that recent advances in multimodal large language models (LLMs), particularly those with vision capabilities, can bridge this gap. While topology optimization has expanded the space of forms that can be computationally explored, LLMs offer a way to connect this search space to human design intent by translating qualitative goals into actionable optimization problems. Together, they could enable designers to directly pursue not only quantitative performance objectives, but also qualitative design goals. Large Language Models (LLMs) such as OpenAI's ChatGPT and Anthropic's Claude have demonstrated strong reasoning across scientific and engineering domains \cite{zhou_large_2022, vaswani_attention_2017, zhao_survey_2023, wei_emergent_2022, touvron_llama_2023} and are increasingly applied in idea generation, formation of hypotheses, and support for analysis \cite{wei_chain--thought_2022, song_evaluating_2025, zhou_hypothesis_2024, zhang_exploring_2025}.

Delegating LLMs the cognitively demanding task of anticipating and reasoning about how coupled parameter decisions affect quantitative performance and qualitative design outcomes relative to designer preferences could fundamentally change how we engage with topology optimization. This parallels the transformation brought about by generative design tools, which enabled designers to evaluate hundreds of diverse alternatives and shifted their attention from low-level parameter and tool management toward higher-level considerations of form and function \cite{loos_towards_2022, saadi_effect_2024, saadi_observations_2023, krish_practical_2011}. However, generative AI still requires expert human involvement during set-up, steering of generated solutions, and careful evaluation of the design space \cite{elrefaie_drivaernet_2025, sung_blendednet_2025}. This motivates our adoption of LLMs to reason over design knowledge and operate at higher levels of abstraction from designer intent.

Recent advances in agentic AI enable us to pursue this goal-oriented reasoning with LLMs. AI agents are autonomous systems in which LLMs serve as reasoning engines that perceive their environment, plan actions, and use external tools through memory-augmented reasoning \cite{stewart_higher-order_2026, nisa_agentic_2025, chowa_language_2025}. Unlike purely reactive LLM generation, agents begin with a goal, decompose it into intermediate steps, and select tool invocations to accomplish each one \cite{yao_react_2022}. Orchestrated multi-agent architectures with specialized roles have shown emergent collaborative behavior \cite{wu_autogen_2023, duan_exploration_2024, wang_agent_2024, stewart_graphagents_2026, ghareeb_robin_2025}, and early engineering-design applications team agents with simulation or code to support selected stages of the design process \cite{elrefaie_ai_2025, ni_mechagents_2023, massoudi_agentic_2025}. Agents have not yet been widely studied in settings where they define their own workflows and carry out end-to-end design without a predefined sequence of steps, which is especially relevant for long-horizon tasks where the appropriate problem formulation is itself unknown in advance.


\subsection{Our Contributions} We introduce TO-Agents, a multi-agent AI framework that autonomously formulates and tunes topology optimization problems to produce optimized designs aligned with human preferences. The designer describes the product design and a desired qualitative quality in natural language, and the system then proceeds without further human input: specialized agents translate that intent into solver settings, run the optimization, visually inspect the resulting structure, evaluate how well it satisfies the request, and revise their parameter choices over successive cycles before preparing the best design for manufacture.

We examine three dimensions:
\begin{itemize}
\item How agents translate qualitative requirements expressed in natural language into TO objectives and parameter selections while navigating tradeoffs among performance, appearance, and other design criteria;
\item How agents improve their decisions through experimentation and collaboration, using prior outcomes and knowledge shared by other agents to better align generated designs with human preferences;
\item How the visual reasoning capabilities of vision-language models can evaluate TO outcomes, relate geometric features to design intent, and provide feedback for iterative learning and refinement.
\end{itemize}
This work positions agents as an interface between designers and topology optimization, bringing engineering performance into closer alignment with human judgment and design intent.

\section{RELATED WORKS}

\begin{figure}[h]
    \centering
    \includegraphics[width=\linewidth]{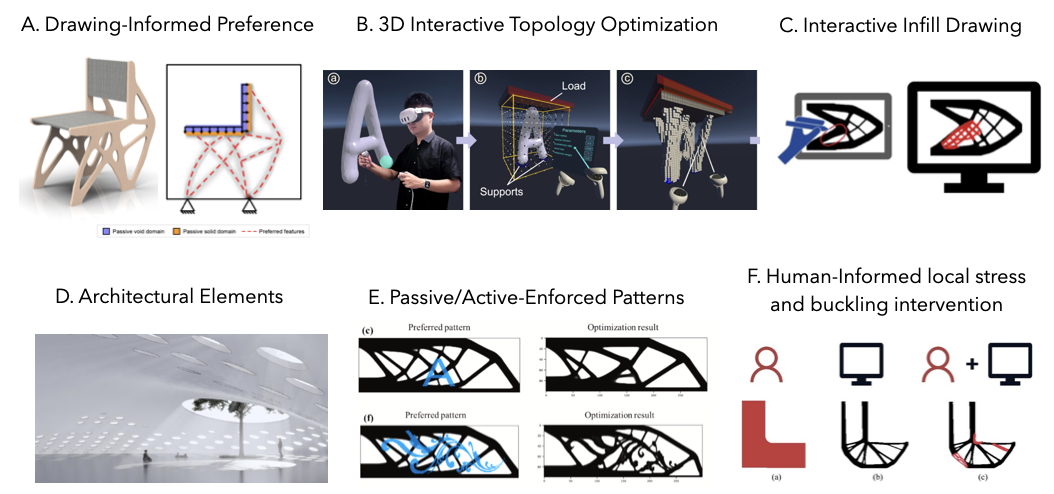}
    \caption{State-of-the-art examples for incorporating human preferences into topology optimization: (A) drawing-informed geometric constraints \cite{inproceedings}; (B) virtual-reality-based specification of three-dimensional preferences \cite{li_interactive_2025}; (C) user-defined infill patterns \cite{schiffer_interactive_2024}; (D) prescribed architectural features, such as holes \cite{meng_direct_2023}; (E) active and passive regions used to enforce design patterns \cite{tian_designer_2026}; and (F) user-directed local feature-size interventions \cite{schiffer_hitop_2023}.}
    \label{fig:preferences}
\end{figure}

In this section, to situate our work within the existing literature, we review prior approaches to incorporating human preferences into topology optimization. An overview is shown in Figure~\ref{fig:preferences}.

\textbf{Pre-processing approach} Yan et al. \cite{yan_detail_2022} developed several extensions to the BESO framework that give designers greater control over the optimization process. MR-BESO enables designers to apply multiple filters with varying sizes across different regions of the design domain. MV-BESO allows the assignment of distinct local volume fractions to individual subdomains according to specific design objectives or personal preferences, leading to more tailored structural layouts. MW-BESO introduces elemental weighting, enabling designers to assign greater importance to selected elements. During sensitivity analysis, these weighted elements receive higher priority, increasing their probability of being preserved in the final topology.

Pre-processing can also incorporate preferences through explicit geometric constraints . For example, designers may prescribe the number and location of holes before optimization. Limiting holes can reduce manufacturing complexity and cost, while strategically introducing them can shape light and shadow to enhance architectural aesthetics \cite{inproceedings, meng_direct_2023}. Preferences may also be built-in through active and passive regions that specify where material must be retained or removed. DPDNTO \cite{tian_designer_2026} demonstrated encoding these regions through a unified interaction parameter and mask-based loss function, enabling neural topology optimization to control geometric features and material distribution according to designer preferences.

\textbf{Post-processing approach} Subjective preferences can also be incorporated into topology optimization in post. One common approach involves manually modifying optimized topologies to enhance their aesthetic qualities. 

One line of research applies principles of aesthetic judgment to refine topology-optimized designs. Although aesthetics are subjective and difficult to formalize, pleasure in design often follows common principles such as unity-in-variety \cite{hekkert2006design}. Variety creates visual interest through differences in form, while unity organizes these differences into a coherent whole. Aesthetic appreciation is generally highest when the two are balanced. This principle has been incorporated into topology optimization by manually refining contours according to Gestalt principles such as similarity, closure, and continuity \cite{loos_towards_2022}.

\textbf{Interactive approach} Interactive topology optimization enables designers to iteratively refine structural solutions through a collaborative feedback loop between human input and optimization algorithms. 

HiTop \cite{ha_human-informed_2023} recognizes that effective design intervention depends on contextual knowledge of a part’s intended use, performance priorities, and manufacturing requirements. Through an interactive interface, users inspect a preliminary topology, identify regions of concern, and locally modify feature-size requirements. These modifications allow users to eliminate, merge, or thicken structural members based on their own judgment. HiTop 2.0 \cite{schiffer_hitop_2023} expands these interventions by allowing users to control the minimum and maximum sizes of both solid and void features in selected regions. The choice of where and how to intervene remains with the user, allowing the optimization to reflect expertise and priorities that are not captured by the compliance objective alone.

Interactive infill topology optimization \cite{schiffer_interactive_2024} allows users to draw their own tailored infill pattern, selecting where it should appear, and controlling how closely the optimized material layout should reproduce it through appearance constraints. The approach has been demonstrated to improve buckling resistance, energy absorption and reduce additive-manufacturing support without introducing complex objectives, additional parameter tuning, or substantial computational costs.

Li et al. \cite{li_interactive_2023} proposed SP-BESO, an interactive subjective-preference topology optimization method that uses random perturbations to generate diverse and unexpected initial topologies. Designers then steer subsequent optimization by scoring these alternatives and providing hand-drawn sketches, which the method converts into weighting factors. Sketches have provided an effective medium for communicating design intent in topology optimization \cite{zhang_machine_2024, feng_sketch2topo_2026, zhu_sketch-guided_2025}. Another BESO-based approach, SBFE-BESO \cite{su_automatic_2024}, uses a user-provided digital image to define the geometry and employs quadtree or octree decomposition to accelerate optimization.

ISP-BESO \cite{li_interactive_2025} combines a similarity constraint with virtual reality to support the interactive exploration of three-dimensional structures. Designers use immersive sculpting and real-time feedback to guide material redistribution, while adjusting the target similarity to balance structural performance with geometric preferences. A museum design case demonstrates the approach’s potential for creating efficient and visually desirable structures.

\section{METHODS}

This section describes the design and implementation of the TO-Agents pipeline. We first present the overall agentic architecture. We then specify the underlying models and libraries used to instantiate each agent. Finally, we describe the tools and memory available to the agents at inference time. 

\subsection{Agentic Architecture}

Our agentic pipeline is composed of a sequence of agents equipped with specialized expertise and tools. Each expert applies their knowledge and responds only when it is their turn to respond, as determined by our predefined logic-based conversation turn architecture. The full agentic architecture is illustrated in Figure~\ref{fig:pipeline}. 

\textbf{Problem Formulation Case Studies} We evaluate our long-horizon agentic design pipeline through two case studies:
\begin{enumerate}
\item A classic cantilever beam problem, commonly used as a benchmark for evaluating new TO methods \cite{sigmund_benchmarking_2022};
\item A phone stand designed to hold a smartphone on a desk, representing a real-world product for which human preferences and aesthetic qualities play a prominent role.
\end{enumerate}
Both case studies use simple boundary conditions and loading scenarios. Their complete problem formulations are provided in Appendix~\ref{app:designer_requests}, Figures~\ref{fig:designer_phone} and~\ref{fig:designer_cantilever}.


\textbf{Initialization} 
To initialize the pipeline, the human designer provides a description of the problem formulation. A Pydantic-based agent interprets the description and transforms it into validated, structured JSON that matches the solver’s expected input format. Pydantic \cite{pydantic2024} is a python-based validation framework linked to the agentic framework. This structured output provides a reliable interface for passing the problem formulation to the topology optimization solver. In Appendix~\ref{app:designer_requests} (Figures~\ref{fig:designer_phone} and~\ref{fig:designer_cantilever}), we assume that the human designer provides a verbose technical natural language description of the problem formulation so that all inputs required for the topology optimization are specified. Future work will investigate how simplifying this description impacts the process.

\textbf{Optimization} 
Next, the topology optimization agent invokes the topology optimization solver, pyFANTOM \cite{ahnobari_pyfantom}, to execute the optimization based on the generated input parameters. We chose this open-source solver because it avoids the black-box constraints of proprietary software, giving us engineers full visibility into the parameters the agent can access and adjust to influence internal algorithmic behavior. With PyFANTOM, the agent can tune parameters throughout a Python-native, scalable, multi-constraint topology optimization framework that supports both structured and unstructured meshes, as well as a wide range of topology optimization objective functions. 

The resulting topology is rendered in 3D, and images of the structure are captured from six orthogonal views (top, bottom, front, back, left, and right) all kept in chat history for subsequent reasoning steps. To improve spatial reasoning clarity, we apply a color mapping with an accompanying colorbar to each voxel. Voxels closer to the camera are rendered in darker colors, while voxels further from the camera appear in lighter colors, as described in Appendix~\ref{app:spatial_perception}. We find that this visualization strategy enables both the human designer and agents utilizing vision–language models to better interpret the topology.

\textbf{Human Preference Feedback} 
Following this, the human designer provides one qualitative feedback describing an aesthetic requirement that the initial topology optimization result fails to satisfy. This is the only point at which the human designer provides input beyond the initial description used to initialize the problem; all subsequent stages of the agentic pipeline proceed fully autonomously. In this study, we have the designer request a structure with increased structural skeletonization and branching complexity promoting a tree-like structure where there exists structural hierarchy and connectivity. 

\textbf{Agentic Deep Reasoning and Refinement} 
This request is passed to the vision agent, which is responsible for implementing the designer’s feedback through a three-stage process. In the first stage, the vision agent analyzes the full chat history, including the previous topology optimization problem parameters and the rendered 3D images of prior structures. In the second stage, the vision agent evaluates the designer’s request alongside the solver parameters available for modification. Based on this information, the vision agent proceeds to the third stage, where it determines which topology optimization parameters should be adjusted and to what extent in order to achieve the designer’s objective. The vision agent is allowed to change any and all of the parameters used to run the optimizer. This multi-stage reasoning workflow is detailed in Appendix~\ref{app:vision_prompt}, Figures~\ref{fig:vision_prompt} and~\ref{fig:vision_revision}.

The vision agent outputs detailed reasoning that identifies which parameters should be changed and explains why those modifications are expected to better satisfy the designer’s intent. Importantly, the agent is not provided with explicit definitions of the topology optimization parameters or any prior guidance about how changing each parameter affects the resulting topology.

Instead, the vision agent relies entirely on knowledge acquired from its foundational training to infer the meaning and relative sensitivity of these parameters when attempting to satisfy the designer’s feedback. In this sense, the multi-objective reasoning process is performed autonomously. The vision agent’s suggestions are returned in natural language and subsequently converted into structured output, which is passed back to the topology optimization agent for a newly revised solver run. The resulting optimized topology is again visually rendered and saved from multiple viewpoints.

\textbf{Agentic Critic Feedback Loop} 
Following this revised optimization, an AI judge agent evaluates both the initial and revised designs. The judge scores the structures based on whether the human designer’s preference criterion of a more branching, intricate structure has been achieved. The scoring criteria and system prompt are detailed in Appendix~\ref{app:judge_prompt}, Figure~\ref{fig:judge_prompt}. The judge must assign a score from 1 to 5 for each design after all revisions have completed. We intentionally design the AI judge to be powered by a larger language model that is trained independently from the vision–language model used in the vision agent. This separation helps promote a more neutral and unbiased evaluation of the outputs.

The judge’s feedback is then added to the conversation history and provided to the vision agent for another revision. At this stage, the vision agent receives a richer context, including the judge’s evaluation, the images of both topology optimization results, and their corresponding solver parameters. As the revision history expands, the vision agent is instructed to use the most recent highest-scoring design in its history as the basis for its next revision while continuing to align with the designer’s request. The maximum number of iterations can be specified by the engineer. For interpretability, we limit the process to five total iterations: one original topology optimization followed by four revision cycles. We demonstrate this over 10 total trial replicates with the exact same system set up for both the cantilever beam and phone stand case study. Figure~\ref{fig:successful_stand} and Figure~\ref{fig:successful_cantilever} show successful designs where the human designer's intent has been met, whereas Figure~\ref{fig:failure_stand} and Figure~\ref{fig:failure_cantilever} show contrasting failure cases. 

\begin{figure*}[t]
    \centering
    \includegraphics[width=1\linewidth]{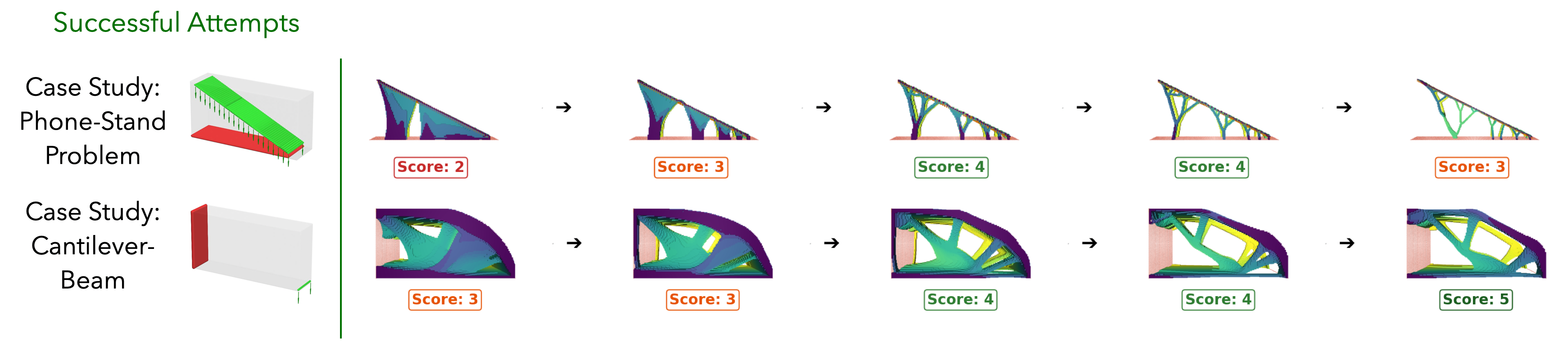}
    \caption{Two case studies illustrating successful multi-turn revision trajectories across the full TOAgents pipeline: phone stand case study (top) and cantilever beam case study (bottom). Both case studies are initialized with the same human preference to increase the hierarchical dendriform visual complexity of the overall topology. AI Judge scores for each successive revision are noted below each design.}
    \label{fig:successful_attempts}
\end{figure*}

\textbf{Production} 
Finally, the highest-scoring structure, as determined by the AI judge, is passed to a manufacturing agent. This agent performs post-processing to convert the topology into a functional phone stand geometry. It adds resolution-scaled functional supports to the optimized density field before converting it to a polygonal mesh via marching cubes and exporting it in OBJ format, as detailed in Appendix~\ref{app:postprocessing}. The final geometry is sent to Bambu Studio \cite{bambustudio2026} for slicing, then 3D printed in standard PLA using a Bambu Lab X1C printer.

\begin{figure}
    \centering
    \includegraphics[width=1\linewidth]{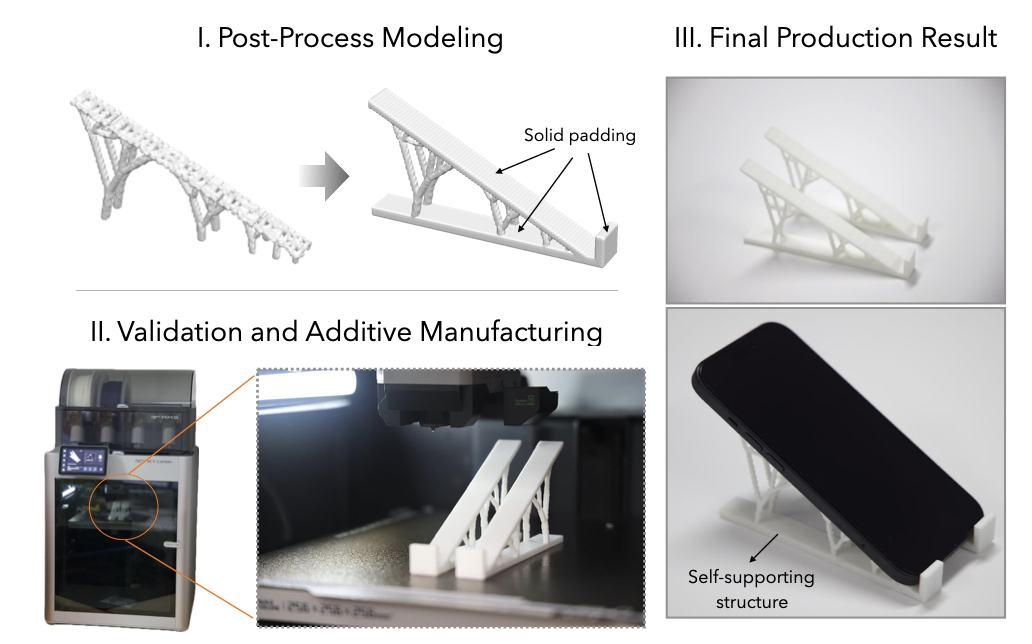}
    \caption{Manufacturing agent prepares the best design made by the agentic system for additive manufacturing which includes post-processing the raw TO structure with solid padding followed by meshing of dense regions. This structure can be directly imported and sliced in a 3D-printing slicer. We verify this pipeline end-to-end with the final print shown in the result.}
    \label{fig:production}
\end{figure}

\subsection{Models and Libraries}

We utilize the open-source AutoGen \cite{wu_autogen_2023} framework to construct our multi-agent architecture, as it represents a state-of-the-art solution for production-grade orchestration of multiple conversational agents.

The models deployed in the system are selected according to the functional role of each agent. The primary model used across the AutoGen framework is the vision–language model Qwen2.5-VL-7B-Instruct \cite{wang_qwen2-vl_2024}. This open-source model was selected for its lightweight 7B parameter footprint and strong benchmark performance in multimodal analysis. Within the pipeline, the model analyzes the 3D structure of topology optimization outputs through the vision agent and generates each agent’s textual responses in the AutoGen chat interface.

To provide a more reliable evaluation stage, we incorporate an AI judge model from a different model family than Qwen \cite{wataoka_self-preference_2024, zheng_judging_2023}. Specifically, we use the open-source Gemma-3-27b-it model \cite{gemma_2025} from Google to assess the results generated by the system. The 27B variant is chosen because its larger parameter size provides stronger evaluation capacity relative to the Qwen model. While the pipeline can also operate with closed-source models such as the Gemini-3 family, this work reports results obtained using the lightweight open-source Gemma-3 model.

All open-source models were served locally using VLLM.ai \cite{kwon_efficient_2023} serving engine.  

\subsection{Tools}

\textbf{Topology Optimization Solvers} We utilize PyFANTOM \cite{ahnobari_pyfantom}, an open-source fast and efficient GPU-ready general package for Finite-Element (FE) based topology optimization. All features are built for general purpose use cases with object oriented setup enabling customization and adaptation to different problem formulations. PyFANTOM serves as the main solver tool call at the disposal of the TO agent.

PyFANTOM solves a SIMP-based minimum-compliance problem, which we summarize here. The design domain is discretized into $n$ elements with element-wise densities $\mathbf{x} = [x_1, x_2, \ldots, x_e, \ldots, x_n]^{\mathrm{T}} \in [0,1]^n$. A linear density filter of radius $r_{\min}$ maps $\mathbf{x}$ to filtered densities $\tilde{\mathbf{x}}$, and each element's stiffness is interpolated as $E(\tilde{x}_i) = E_{\min} + \tilde{x}_i^{\,p}(E_0 - E_{\min})$, where $E_0$ and $E_{\min}$ are the solid and void Young's moduli and $p$ is the SIMP penalty exponent. With Poisson's ratio $\nu$ fixed, the assembled global displacements $\mathbf{U}(\tilde{\mathbf{x}})$ satisfy the FE equilibrium $\mathbf{K}(\tilde{\mathbf{x}})\mathbf{U}(\tilde{\mathbf{x}}) = \mathbf{F}$ for the applied load vector $\mathbf{F}$. The optimization problem is
\begin{equation}
\label{eq:to_problem}
\begin{aligned}
\text{find} \quad & \mathbf{x} = [x_1, x_2, \ldots, x_e, \ldots, x_n]^{\mathrm{T}} \\
\text{minimize} \quad & c(\tilde{\mathbf{x}}) = \mathbf{F}^{\mathrm{T}}\mathbf{U}(\tilde{\mathbf{x}}) \\
\text{subject to} \quad & v(\tilde{\mathbf{x}}) = \tilde{\mathbf{x}}^{\mathrm{T}}\mathbf{v} - \bar{v} \le 0 \\
& \mathbf{x} \in \mathcal{X}, \quad \mathcal{X} = \{\mathbf{x} \in \mathbb{R}^n : \mathbf{0} \le \mathbf{x} \le \mathbf{1}\},
\end{aligned}
\end{equation}
where $c$ is the structural compliance, $\mathbf{v}$ is the vector of element volumes, and $\bar{v} = f \, V_{\mathrm{tot}}$ is the volume budget set by the volume fraction $f$. The compliance sensitivity used by the gradient-based optimizer takes the standard SIMP form
\begin{equation}
\label{eq:compliance_sens}
\frac{\partial c(\tilde{\mathbf{x}})}{\partial \tilde{x}_i}
  = \mathbf{u}_{di}(\tilde{\mathbf{x}})^{\mathrm{T}} \left[\, p \tilde{x}_i^{\,p-1} (E_0 - E_{\min}) \mathbf{k}_i^0 \,\right] \mathbf{u}_i(\tilde{\mathbf{x}}),
\end{equation}
where $\mathbf{k}_i^0$ is the element stiffness matrix evaluated at unit modulus, and $\mathbf{u}_i$ and $\mathbf{u}_{di}$ are the displacement and adjoint displacement vectors restricted to element $i$.

\section{RESULTS AND DISCUSSION}

The agentic pipeline reveals collective behavior resulting from the autonomy granted to the agents in their interpersonal interactions. The agents are therefore able to influence one another without human intervention, experimentally moving the system closer to the human designer’s intended objective, with optimization simulations and tool invocations serving as validation and feedback.

\begin{figure*}[t]
    \centering
    \includegraphics[width=1\linewidth]{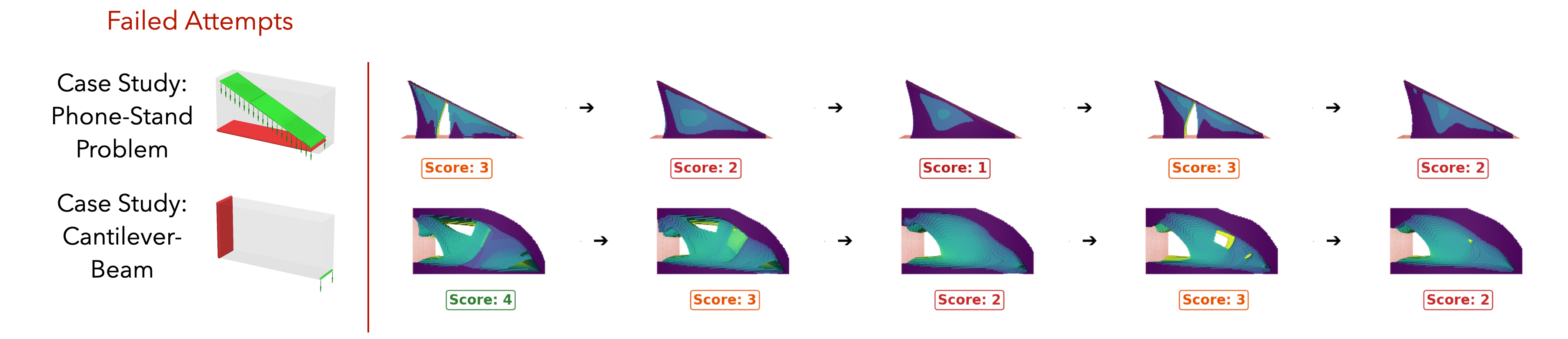}
    \caption{Two case studies illustrating failed multi-turn revision trajectories across the full TOAgents pipeline: phone stand case study (top) and cantilever beam case study (bottom). Both case studies are initialized with the same human preference to increase the hierarchical dendriform visual complexity of the overall topology. AI Judge scores for each successive revision are noted below each design.}
    \label{fig:failed_attempts}
\end{figure*}

\subsection{History-Conditioned Adaptation}
\label{sec:history_adaptation}

Across our experimental replicates, we consistently observe a cycle of refinements in which the vision agent uses the growing history of prior designs, parameter settings, and AI-judge feedback to progressively learn how to manipulate TO parameters toward the human designer's qualitative objective. We first illustrate this behavior with a single representative replicate, then aggregate over all replicates in both case studies to characterize which parameters the agents actually use as their levers. An exemplar set of successful runs enabled by history-conditioned adaptation for both case studies is shown in \ref{fig:successful_attempts}.

\subsubsection{Numerical Example}

We take Figure~\ref{fig:successful_stand} replicate B as a running example. Table~\ref{tab:summary_auto_5turn3} summarizes how the agent moves the parameters across the four revisions, and Figure~\ref{fig:successful_stand} shows the corresponding structures. The full vision-agent reasoning and AI-judge evaluation for each turn can be found in the supplementary material; here we extract only notable highlights in their reasoning.

\paragraph{Identifying the Right Levers.}

After viewing the structure, the vision agent judged the baseline original design as too plate-like. 
\begin{agentquote}{Vision Agent}
The structure appears relatively flat and lacks significant branching. There are some thin members but they do not form a complex network. The overall shape suggests it might be more plate-like rather than truss-like.
\end{agentquote}

The vision agent immediately identifies the volume fraction $f$ and SIMP penalty $p$ as the two parameters most likely to promote a thinner, more branched structure, and respects the lower bound on the filter radius $r_{\min}$:
\begin{agentquote}{Vision Agent}
Lowering the volume fraction increases the amount of void space, which encourages the optimizer to create thinner members \dots A higher penalty parameter penalizes thicker structures more heavily \dots The filter radius should not be reduced further as it could lead to overly fine elements that might not be practically feasible.
\end{agentquote}

Accordingly, it drops $f$ from $0.15$ to $0.10$ and raises $p$ from $3$ to $5$. The AI judge scores the resulting design ($s=4$) above the original ($s=3$) and explicitly attributes the increase to the appearance of richer hierarchy. This appeal to well-defined arches and supports confirms that the multimodal channel is being used. 

\paragraph{Doubling Down on the Working Strategy.}

Looking at the revised design, the vision agent observes progress but is not satisfied: 

\begin{agentquote}{Vision Agent}
The current structure appears to have some branching but lacks significant thinness and complexity. It still has areas that look like thick plates, and there isn't enough internal hierarchy or connectivity.

\textbf{1.\ Thinning Members:} The structure is not sufficiently thin, which is crucial for achieving a skeletal design.\\
\textbf{2.\ Internal Load Paths:} The structure seems to rely heavily on a single dominant load path, lacking multiple internal load paths.\\
\textbf{3.\ Connectivity:} There is room for increased internal connectivity to promote a more truss-like behavior.
\end{agentquote}

With two designs in its context, the vision agent observes that thick columns still dominate revision 1 and pushes the same two levers further: $p$ from $5\!\to\!7$ and $f$ reduced again to $0.05$. Revision 2 exhibits clear hierarchical branching and the judge scores it $0.5$ points higher than revision 1, although with $5\%$ lower confidence as the comparison becomes finer. Interestingly, the agent also claims that the optimizer settings seem fine, so no changes are necessary and opts to focus its targeted adjustments only to TO hyperparameters. 

In revision 3 the agent acknowledges that the previous revision was already complex but pushes for more improvement. The agent now diversifies its strategy, increasing the mesh resolution $(n_{elx}\times n_{ely}\times n_{elz})$ from $128\!\times\!64\!\times\!16$ to $160\!\times\!80\!\times\!20$ and tightening the optimizer function tolerance, producing some small void regions but no large global change; the judge correctly assigns it the same score as revision 2. One faulty behavior is that the agent can give mismatched instructions, an artifact of model hallucination. Specifically, the agent prose suggests increasing function tolerance, but the suggested numerical values suggest a decrease. 

\begin{agentquote}{Vision Agent}
Higher mesh resolution can help capture finer details and promote more complex branching.
\end{agentquote}

\begin{agentquote}{Vision Agent}
PGD optimizer might benefit from slight adjustments to improve convergence towards a more skeletal structure\dots Increase the \texttt{fun\_tol} slightly to allow for more exploration (from 0.0001 to 0.00005).
\end{agentquote}

The judge affirms, based on visual inspection, that the original is the most simple with only a few dominant load paths and that higher scores were given to these later revisions for achieving a richer network of load paths.

\paragraph{Overshooting.}
With these successes in store, the agent continues to push in the same direction with increasing $p$ to $12$, $f$ to $0.02$ and mesh further refined. The consequence of these changes are that there is so little material afforded that the optimizer can only afford a few primary members and thus the design fails to improve. The judge now scores the result \emph{below} revisions 2 and 3, identifying that the parameters have been pushed too far. 

From our perspective, the design still satisfies the qualitative objective, and the judge's confidence is notably lower in this evaluation; the run ends before any recovery from the overshoot is possible. Even so, the trajectory as a whole moves clearly toward the human designer's intent, and we treat this replicate as a representative successful run.

\begin{table}[htbp]
\centering
\caption{Parameter trajectory across revisions for Figure~\ref{fig:successful_stand} replicate B.}
\label{tab:summary_auto_5turn3}
\small
\resizebox{\columnwidth}{!}{%
\begin{tabular}{l r r r r r}
\toprule
\textbf{Parameter} & \textbf{Original} & \textbf{Rev.\ 1} & \textbf{Rev.\ 2} & \textbf{Rev.\ 3} & \textbf{Rev.\ 4} \\
\midrule
Filter radius, $r_{\min}$ & 2.00 & 2.00 & 2.00 & 2.00 & 2.20 \\
$n_{elx}$ & 128 & 128 & 128 & 160 & 192 \\
$n_{ely}$ & 64 & 64 & 64 & 80 & 96 \\
$n_{elz}$ & 16 & 16 & 16 & 20 & 24 \\
Opt.\ fun.\ tol. & $10^{-4}$ & $10^{-4}$ & $10^{-4}$ & $5\!\times\!10^{-5}$ & $10^{-5}$ \\
SIMP penalty, $p$ & 3.0 & 5.0 & 7.0 & 10.0 & 12.0 \\
Volume fraction, $f$ & 0.150 & 0.100 & 0.050 & 0.030 & 0.020 \\
\bottomrule
\end{tabular}
}
\end{table}

Three behaviors highlighted by this replicate recur across the dataset: (i) the agent discovers a small set of effective levers within the first revision; (ii) it amplifies a strategy that the judge has rewarded; and (iii) it eventually overshoots, providing the judge with a clear negative signal. We next show that these behaviors generalize across replicates and case studies.

\subsubsection{Aggregate Parameter Usage}

To quantify which parameters the agents actually exercise, we compare every revision of all $10$ replicates to its original baseline and report, for each parameter, the average signed change, the average absolute change, and the count of revisions in which it was modified. Results for both case studies are summarized in Table~\ref{tab:top_params_combined}.

\begin{table*}[htbp]
\centering
\caption{Top parameters modified by the agents across both case studies (across $40$ revisions of $10$ replicates each). Set A corresponds to the phone stand problem and Set B to the cantilever beam problem. ``n/m'' indicates the parameter was available to the agent but never modified in that case study.}
\label{tab:top_params_combined}
\small
\begin{tabular}{l r r r r r r}
\toprule
 & \multicolumn{3}{c}{\textbf{Set A Case Study}} & \multicolumn{3}{c}{\textbf{Set B Case Study}} \\
\cmidrule(lr){2-4} \cmidrule(lr){5-7}
\textbf{Parameter} & \textbf{Avg.\ $\Delta$} & \textbf{Avg.\ $|\Delta|$} & \textbf{Count} & \textbf{Avg.\ $\Delta$} & \textbf{Avg.\ $|\Delta|$} & \textbf{Count} \\
\midrule
SIMP penalty, $p$         & $+4.81$    & $4.81$    & 40  & $+3.75$    & $3.86$    & 36  \\
$n_{elx}$                 & $+131.8$   & $132.7$   & 34  & n/m        & n/m       & n/m \\
$n_{elz}$                 & $+16.5$    & $16.6$    & 34  & n/m        & n/m       & n/m \\
$n_{ely}$                 & $+68.0$    & $68.4$    & 33  & n/m        & n/m       & n/m \\
Filter radius, $r_{\min}$ & $+0.09$    & $0.68$    & 32  & $+0.36$    & $0.64$    & 20  \\
Volume fraction, $f$      & $-0.03$    & $0.08$    & 32  & $-0.10$    & $0.12$    & 35  \\
Heaviside, $\beta$        & n/m        & n/m       & n/m & $-1.00$    & $1.00$    & 15  \\
Multigrid tolerance       & $-10^{-4}$ & $10^{-4}$ & 13  & n/m        & n/m       & n/m \\
Number of iterations      & $+91.7$    & $91.7$    & 6   & $+87.5$    & $87.5$    & 10  \\
Young's modulus, $E_0$    & $-0.32$    & $0.38$    & 6   & n/m        & n/m       & n/m \\
Poisson ratio, $\nu$      & $-0.05$    & $0.05$    & 4   & n/m        & n/m       & n/m \\
Opt.\ fun.\ tol.          & n/m        & n/m       & n/m & $-10^{-4}$ & $10^{-4}$ & 2   \\
\bottomrule
\end{tabular}
\end{table*}

In the phone-stand study, the SIMP penalty $p$ is the single most reliable lever: it is increased in every one of the $40$ revisions, by an average of $+4.8$. Mesh refinement is the next dominant pattern with $n_{elx}$, $n_{ely}$, and $n_{elz}$ are each modified in $33$--$34$ revisions, and the near-equality of signed and absolute averages shows the agents almost exclusively \emph{increase} resolution rather than coarsen. The volume fraction $f$ is touched as often (32 revisions) but with much smaller magnitude (avg $|\Delta|=0.08$), consistent with fine-tuning rather than exploration. The filter radius $r_{\min}$ shows the opposite signature: a similar count, but a near-zero signed average against a much larger absolute average ($0.09$ vs.\ $0.68$), indicating that the agents lack a single consistent direction here and instead push it both ways depending on context. Material properties ($E_0$, $\nu$) and the iteration budget are reserved for a small number of revisions, suggesting the agents recognize them as more drastic levers.

The cantilever-beam study reveals a markedly more focused strategy: only six numeric parameters are modified across all $40$ revisions, and mesh resolution and material properties are never touched. The SIMP penalty $p$ is again the dominant lever ($36$ revisions, $+3.75$ on average), and the volume fraction $f$ is consistently \emph{reduced} ($35$ revisions, signed average $-0.10$). The Heaviside projection toggle $\beta$ is exercised in every case where it is touched ($15$ revisions, signed and absolute averages both $1.0$), functioning as a binary switch the agents flip on or off. We attribute the narrower lever set to the more constrained problem formulation: the cantilever uses a single point load, whereas the phone stand applies multiple loads along a diagonal that admit more distributed load paths and therefore reward exploration of mesh and material parameters. We elaborate on the qualitative behavior of the cantilever replicates in the supplementary material; here it suffices that history-conditioned adaptation is observed in both studies, but with study-specific parameter strategies that the agents discover from the feedback signal alone.

\color{black}

\subsection{Observed Recovery from Low-Scoring Revisions} 

\begin{figure}
    \centering
    \includegraphics[width=1\linewidth]{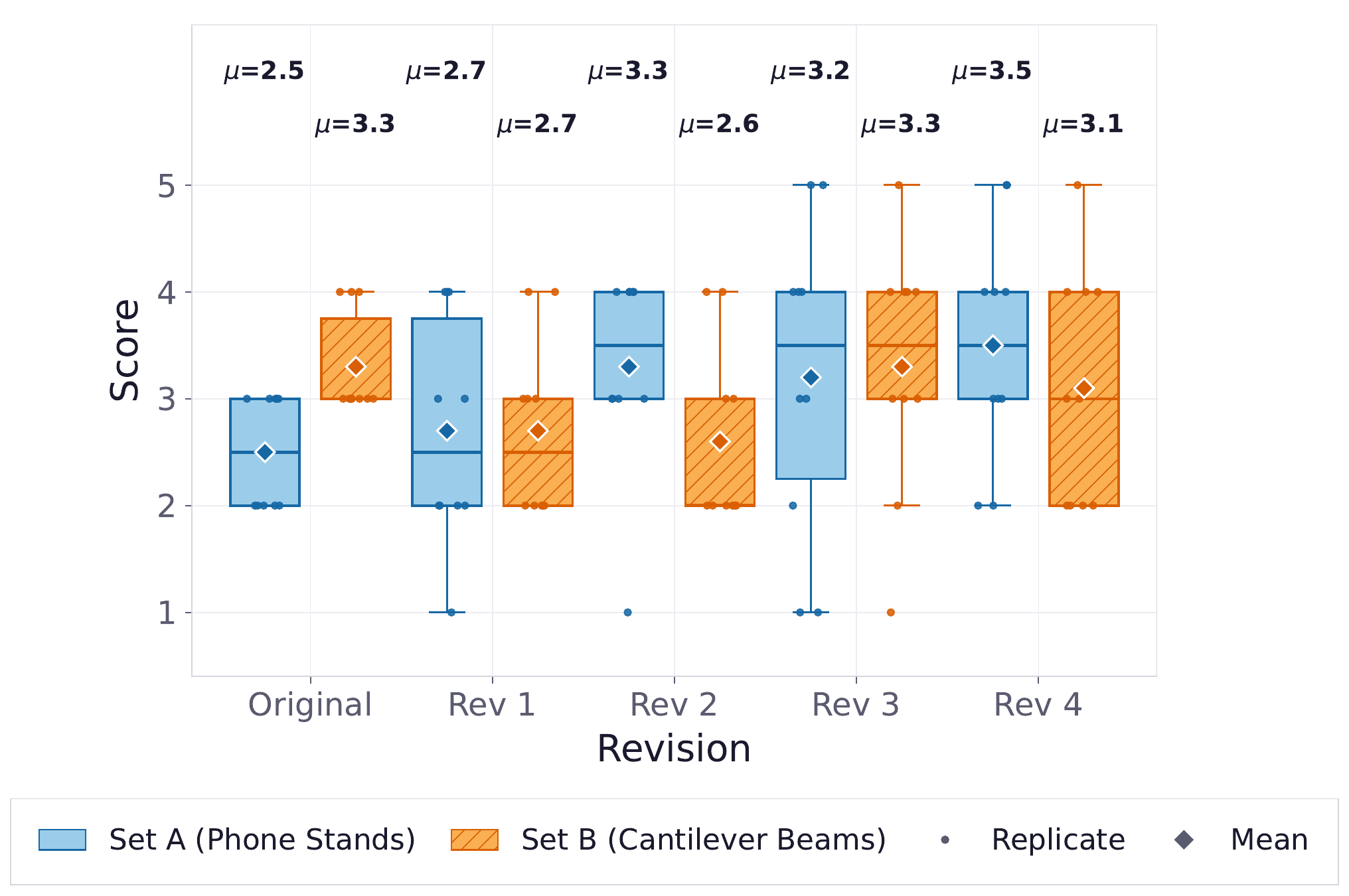}
    \caption{AI judge scores across 10 replicates per revision where set A (red solid) represents the phone stand case study and set B (orange hatched) represents the cantilever beam case study. Set A shows both an upward shift in central tendency (\(\mu: 2.5 \rightarrow 3.5\) across revisions) and a broadening distribution toward higher scores. Set B's mean remains roughly stable near \(\mu \approx 3.0\), but its distribution widens comparably, expanding from a tight initial cluster around 3 to span the full 2--5 range by Revision~4.}
    \label{fig:average_scores}
\end{figure}

Another behavior observed in this agentic system is its capacity for recovery from  low scoring revisions, enabling the agents to detect and adapt to their own errors and recover when their decisions diverge from the intended objective. In these cases, the agents leverage the history of prior revisions to reason backward and adapt their decisions in response to critical feedback from the AI judge, weighing its previous successful and unsuccessful choices. By identifying the most effective prior revision, the agents adjust the relevant parameters and continue the process from that point, guiding the optimization back toward the desired trajectory. 

A prime example of this is shown in Figure~\ref{fig:successful_stand} replicate C. In the first revision, the agent demonstrates strong intuition in selecting parameter adjustments that promote the desired branching structure which yields a high score over the original design. However, in revisions 2 and 3, the degree of branching progressively decreases. This is reflected in the AI judge's assessment after revision 3 was completed: 

\begin{llmbox}[title=AI Judge Agent Evaluation]

{\footnotesize\bfseries Comparative Analysis:}

Here's an evaluation of the aesthetic quality of the four topology optimization results, focusing on global structural complexity.

\textbf{1.\ Image A (Original): Score -- 2}\\
\textit{Justification:} The structure exhibits a very simple global form. It consists of a single dominant arch and a minimal number of supporting members. There is little branching and almost no hierarchical organization beyond the primary load path.\\
\textit{Confidence: 95\%}

\textbf{2.\ Image B (Revision 1): Score -- 4}\\
\textit{Justification:} This design demonstrates a significantly higher level of global complexity. Multiple arches and branching supports form a rich hierarchical structure. The network of load-carrying members is dense and interconnected, providing several possible load paths.\\
\textit{Confidence: 90\%}

\textbf{3.\ Image C (Revision 2): Score -- 3}\\
\textit{Justification:} Image C presents a moderate level of complexity. It contains a primary arch with secondary branching supports, but the branching network is less extensive and the hierarchy less developed than in Image B. The structure is more intricate than Image A but does not reach the complexity of Image B.\\
\textit{Confidence: 85\%}

\textbf{4.\ Image D (Revision 3): Score -- 3}\\
\textit{Justification:} Similar to Image C, this design exhibits moderate global complexity. It features a single arch with branching supports. While the form appears slightly more refined, the overall level of branching and hierarchy is comparable to Image C.\\
\textit{Confidence: 80\%}

{\footnotesize\color{black!60}\hrulefill}

\textbf{Summary:}
Image B stands out as the most globally complex design, characterized by a dense network of interconnected members and a clear hierarchical structure. Images C and D exhibit comparable levels of moderate complexity, while Image A represents the simplest structure. The scoring reflects this progression in global structural richness.

\end{llmbox}

Tracing the revision history reveals a foundational error that contaminates the agent's subsequent credit assignment. At revision 1, the agent made two mechanically sound moves, reducing the volume fraction and increasing the SIMP penalty, but simultaneously increased the filter radius under the mistaken belief that a larger $r_{\min}$ would promote thinner members. A human expert in TO intuitively knows that the opposite is true since a larger filter radius enforces a larger minimum feature size and therefore produces thicker, smoother members.

\begin{agentquote}{Vision Agent}
A larger $r_{\min}$ value will allow for the removal of smaller, less critical elements, promoting the formation of a more skeletal structure.
\end{agentquote}

Given that the first revision achieved a high score, it assumed that including this choice was a good decision. However, further increasing the filter radius in the same incorrect direction led to an observed reduction in score. In the subsequent final revision, the vision agent finally recognizes the error based on the optimization history. It attributes the decline in performance to its earlier faulty decision to increase $r_{\min}$. Accordingly, the vision agent reverses its prior reasoning and recommends drastically reducing $r_{\min}$ from $r_{\min}=3.75$ to $r_{\min}=2.5$.

The agent was able to arrive at this adjustment by revisiting revision 1, which holds the highest score in the revision history, and using it as a reference point for subsequent parameter refinement. By building from this earlier successful state and implementing these suggestions, the vision agent helps the system recover and produce a high-scoring design for the final revision. 

The adjustments made across this revision can be seen in Table~\ref{tab:summary_auto_5turn4}. 

\begin{table}[htbp]
\centering
\caption{Observed recovery in parameter values across revisions for Figure~\ref{fig:successful_stand} replicate C.}
\label{tab:summary_auto_5turn4}
\small
\resizebox{\columnwidth}{!}{%
\begin{tabular}{l r r r r r}
\toprule
\textbf{Parameter} & \textbf{Original} & \textbf{Revision 1} & \textbf{Revision 2} & \textbf{Revision 3} & \textbf{Revision 4} \\
\midrule
Filter radius, $r_{\min}$ & 2 & 3 & 3.5000 & 3.7500 & 2.5000 \\
$n_{elx}$ & 128 & 128 & 160 & 256 & 512 \\
$n_{ely}$ & 64 & 64 & 80 & 128 & 256 \\
$n_{elz}$ & 16 & 16 & 20 & 32 & 64 \\
Multigrid tolerance & 0.00010 & 0.00005 & 0.00001 & 0.00005 & 0.00005 \\
Optimizer change tolerance & -- & -- & 0.00010 & 0.00005 & 0.00010 \\
Optimizer function tolerance & 0.00010 & 0.00010 & 0.00010 & 0.00005 & 0.00010 \\
SIMP penalty, $p$ & 3 & 4 & 5 & 6 & 8 \\
Volume fraction, $f$ & 0.1500 & 0.1000 & 0.1500 & 0.1200 & 0.0500 \\
\bottomrule
\end{tabular}
}
\end{table}

This recovery behavior is also evident in Figure~\ref{fig:successful_stand}, replicate E. In this case, the first few revisions successfully improved the score relative to the baseline, but performance plateaued at $s=4$ in revisions 1 and 2. This plateau placed pressure on the agent to take more drastic actions by adjusting multiple levers simultaneously in revision 3. Here, the agent wrestles with two competing interpretations of the objective: increasing structural branching, which requires additional material, while producing thinner members which requires reducing material usage. The agent chose to simultaneously raise $f$ from 0.05 to 0.15, drop $r_{\min}$ from 1.5 to 1.0, refine the mesh from $256{\times}128{\times}32$ to $512{\times}256{\times}64$, and halve $E$, among other changes.

The result was a failure state for revision 3 in which branching disappeared entirely and the AI judge correctly assigns this design a score of $s=1$, describing the structure as "almost a monolithic shell." There was simply too much material available and not enough filter-enforced minimum feature size, to force the optimizer toward a branching topology. 

In the final revision, the agent is able to recover by returning to a prior more successful configuration, which enables it to fix its reasoning that reducing $f$ produces more favorable results. The resulting design introduces even greater branching than the earlier designs, ultimately producing a structure that strongly satisfies the human designer’s objective.

For the cantilever beam problem, recovery behavior is likewise evident across several replicates. For example, in Figure~\ref{fig:successful_cantilever}, replicate E, the first few revisions produced excessively thick branches. Consequently, by revision 3, the agent effectively reverted to the original design and corrected its earlier parameter adjustments, ultimately achieving a record-high score of $s=4$.

In some cases, however, progress is non-monotonic. Intermediary revisions can outperform the final revision, indicating that the agent is not always able to fully build on earlier successes. For example, in Figure~\ref{fig:successful_stand}, replicate A, revision 3 receives a higher score than the final revision 4. Similarly, in Figure~\ref{fig:successful_stand} replicate D, revision 1 clearly represents the best-performing design. Despite this, we still consider these runs successful, as at least one design within the evolution of revisions achieves the human designer’s vision for the objective. This is evident also in the case for the cantilever beam study for instance in Figure~\ref{fig:successful_cantilever} replicate D.

\subsection{Trajectory-Level Exploration}

Earlier, we observed that the agents discovered that focusing on a small set of optimization parameters, including the SIMP penalty $p$ and the volume fraction $f$, was sufficient to achieve the objective. However, emergent behavior arises when the agent begins to reason over the design space more broadly. This shift is evident when the agent starts to consider additional factors such as material properties, boundary condition placement, load placement, and the optimizer selection itself, effectively altering the problem as a whole in order to better achieve the human designer’s envisioned outcome.

In Figure~\ref{fig:successful_stand} replicate E, the agent reaches beyond the standard optimization settings and attempts to reason about how material properties interact with the TO solver. In revision 2, it leaves the Young's modulus $E_0$ at $1.0$ and reduces the Poisson ratio to $\nu = 0.25$. By revision 4, it halves the Young's modulus to $E_0 = 0.5$ and restores $\nu = 0.3$, justifying the change with the claim that a lower $E_0$ "promotes thinner members."

This reasoning is mechanistically incorrect for the problem at hand. From the compliance sensitivity in Eq.~\eqref{eq:compliance_sens}, $E_0$ enters as a uniform scalar multiplier on every element sensitivity. Scaling $E_0$ rescales the gradient by the same constant at every design variable, leaving the ranking of elements and therefore the optimal density field unchanged. Halving $E_0$ would double the reported compliance values but produce an identical topology. Poisson's ratio is the only material property that genuinely affects the optimum: $\nu$ enters the elasticity tensor non-uniformly through the $1/[(1+\nu)(1-2\nu)]$ prefactor and the differing scaling of normal and shear blocks, so element sensitivities shift non-uniformly across the domain.

Furthermore, in this replicate, the vision agent maintains the existing boundary conditions to preserve the structural functionality of the phone stand. This behavior suggests that the agent recognizes how the problem setup relates to the functional role of the design. In particular, it is cautious about modifying the diagonal face, which serves as the primary contact surface for the phone. The vision agent's reasoning on this point in revision 2 reads:

\begin{agentquote}{Vision Agent}
\textbf{Boundary Conditions.} Boundary conditions can affect how the structure responds to loads. Ensure the boundary conditions are correctly applied to maintain functionality while allowing for more complex deformation patterns.
\end{agentquote}

At another meta-thinking level, agents can reason over the selected optimization algorithm. For example, in Figure~\ref{fig:failure_stand}, replicate D, the vision agent experiments with different optimization strategies, alternating between the Method of Moving Asymptotes (MMA) optimizer and the Projected Gradient Descent (PGD) optimizer, which can produce substantially different visual outcomes.

In the first revision, the vision agent suggests applying an alternative optimizer in place of PGD. The vision agent suggests using plain gradient descent (GD) instead. However, since plain GD was not available to the agent as a tool, the TO agent proceeded with the MMA optimizer as a more robust alternative to PGD, while keeping to the remaining parameter adjustments as suggested. This design was punished by the AI judge for resulting in an overly simplistic design with little branching hierarchy. 

Thus, the vision agent returns to the PGD optimizer in the second revision. In revision 3, the vision agent calls for the use of MMA explicitly despite previous failures with this algorithm, but this time with a much lower volume fraction $f$ applied to increase void space based on its learnings from the success of revision 2. As a result, this third revision is the highest scoring in the sequence. 

Although the manufactured design is not classified by the human designer as fully meeting the objective (placing this replicate in the failure category overall), the revision 3 result demonstrates that the agent can explore a method using a completely different algorithm that can yield a uniquely rich branching that the judge values highly. With additional revision cycles, this productive direction could potentially have been refined further.

A similar pattern appears in the cantilever beam case study. In Figure~\ref{fig:successful_cantilever}, replicate A, the agent applies MMA during the first three revisions but switches back to PGD by revision 4 after recognizing that MMA was not providing sufficient improvement. The agent’s output shows that it correctly identifies the current optimizer as MMA based on the history, explicitly names PGD as an alternative, and interprets the switch from MMA to PGD as evidence that it is tracking its own optimizer state across revisions.

\begin{agentquote}{Vision Agent}
Optimizer Type: Switching from MMA to PGD might provide a more robust exploration of the design space, leading to a more complex and intricate structure...

While the MMA optimizer was effective in Revision 3, exploring other optimizers like PGD might yield different results.
\end{agentquote}

This suggests that the agent can determine through iterative experimentation when switching optimizers is beneficial for aligning the design more closely with the intended objective.

\subsection{Failure Cases}

This pipeline reveals several weaknesses of the agents when performing a multi-objective, long-horizon task, particularly due to the stochastic nature of the responses and the inconsistent retrieval of information from memory.

\subsubsection{Meeting Human Preferences}
There are cases in which the agent struggles to produce designs that align with the human designer’s vision. Sometimes agents double down on a bad strategy, thinking they had not pushed it far enough but this only causes further stray from the desired final form. A representative failure of this kind appears in Figure~\ref{fig:failed_attempts}: in the initial revision the agent increased the volume fraction, and in subsequent revisions it continued in the same direction, expecting that adding more material would spread into a branch-like structure. Other selected replicates are shown as a collection in Figure~\ref{fig:failure_stand} for the phone stand study where the designs do not meet the human designer's mental model. 

In other cases, all of the mechanistic changes made to the parameters are sound but the final structure does not meet the human rubric for an aesthetically successful design even when the AI Judge concludes that it does. In Figure~\ref{fig:failure_stand}, replicate C, the final structure appears somewhat disordered and does not exhibit the organized hierarchical branching envisioned by the human designer that other runs were better able to achieve. The agent however still exhibits good decision-making: it increases the optimizer’s change tolerance while decreasing the function tolerance to encourage broader exploration of the design space. As a result, the final revision does appear the most branch-like and is judged by the AI judge favorably among the generated designs receiving the highest score.

\subsubsection{Use of Available Tools}
Other times the agent makes suggestions that are not within the tool-set. In Figure~\ref{fig:failure_stand}, replicate B, the agent attempts to introduce a continuation schedule on the SIMP penalty $p$. However, the system did not provide the agent with the capability to implement this function, leading it to attempt to invoke a feature that was not available to it. As a result, the agent neither learned nor recognized later in review of the history that the intended functionality was not applied. Even still, the agent’s intent was sound: it proposed gradually increasing $p$ in order to penalize compliance more effectively without overly stiffening the structure too quickly. The agent had hoped to introduce a $p$-continuation schedule that penalizes thick regions more heavily over time. Without this ability, the optimization results shown in this figure do not accurately reflect the agent’s intended strategy and appear as failures with little to no branching. As a result, the agent remains unaware and cannot realize this limitation on its own. Implementing a mechanism to detect such unsupported functionality is left for future work.

Figure~\ref{fig:failure_cantilever} shows the replicates in the cantilever beam case study for which none of the generated designs satisfy the human designer’s requested qualitative specifications. In Figure~\ref{fig:failure_cantilever} replicate A, the agent does not recognize until the final revision that reducing the volume fraction $f$ produces a more skeletal, branching structure that is evaluated more favorably by the AI judge. This case is therefore considered a failure because the reduction occurs too late, and not aggressively enough, to achieve the desired result within the allotted number of revisions, even though the agent may have succeeded if additional revision cycles had been allowed.

\subsubsection{Selective Memory}

The agent does not reliably adhere to explicit constraints. The rule stated: 

\begin{quote}
``Do \textbf{NOT} go below 1.5 in density filter radius,'' 
\end{quote}

The agent nevertheless recommends reducing the filter radius to increase visible branching detail. While the intuition behind adjusting the parameter is reasonable, the behavior demonstrates that the agent may ignore or override constraints when it believes doing so will help it achieve its goal. More generally, this suggests that instructions are not followed consistently and that attention to constraints may be selective.

In Figure~\ref{fig:successful_cantilever}, replicate E, the agent recommends $r_{\min} = 1.0$ in revision 1 and $r_{\min} = 0.5$ in revision 2 without acknowledging the imposed constraint. By the third revision, however, the agent identifies the rule and explicitly requests adherence to the minimum radius constraint of $r_{\min} = 1.5$.

Furthermore, agents engage with available information selectively. The agent reliably retains forward-pointing facts (current state, goal, generic claims) but often loses backward-pointing facts (what worked, what didn't, why). Access to conversation history does not mean that the agent exploits it as a useful tool for its reasoning in order to better trace score deltas back to specific parameter changes. The agent sometimes will directly claim in its response that it is reasoning based on insights gained from previous iterations, but will proceed to report key observations that are generic mechanistic claims without directly referencing previous designs. The agent's selective memory is partially compensated by deterministic structural memory from the rebase to the best scoring design demonstrating how the engineered architecture of the set up can be a useful guardrail.

\subsubsection{Physical Intuition}

As discussed in prior sections, the agents can form internally plausible but mechanistically incorrect explanations. For example, the agent may reason that increasing the volume fraction should enable more branching by allowing more material to be used, when in practice this can otherwise move the design further from the objective where reducing volume fraction would in fact more easily promote thin branching. 

Another failure mode is that agents conflate doing more work with being more productive. If scores are unable to increase after several revisions, the agent will escalate the situation by bundling multiple parameter changes at once. In Figure~\ref{fig:successful_stand} replicate E, after two successive revisions with scores of 4, the agent calls for six simultaneous parameter changes in the third revision. While each change was defensible, the bundle caused a catastrophic collapse in score ($s=1$), defeating credit assignment and making it difficult for agents to track which levers helped and which hurt.

\subsection{AI Judge Scoring Trends}

The AI judge scores across replicates shown in Figure~\ref{fig:average_scores} reveal that early revisions display relatively constrained performance, with mean scores largely clustered between 2 and 4. In later revisions, the agent begins to explore a broader design space: the distribution of scores expands substantially, with outcomes ranging from roughly 1 to 5 by the third revision for both sets.

For Set A, corresponding to the phone stand case study, the mean scores increase steadily across revisions. In contrast, Set B, representing the cantilever beam case study, exhibits greater variability, with scores declining in the early revisions before partially recovering, though the final mean remains slightly below the initial value. This difference may be partly explained by the higher starting performance in Set B (3.3) compared with Set A (2.5), leaving less room for measurable improvement. As a result, Set A appears to benefit more clearly from the iterative revision process, showing a consistent upward trajectory in performance.

This widening spread in scores by late revisions supports the observation that the agent increasingly explores diverse design solutions as revisions progress. For Set A, several replicates improve steadily across revisions, contributing to the rising mean. In Set B, replicate trajectories are more heterogeneous: some runs improve substantially in later revisions, while others stagnate or decline. These results suggest that the revision process enables exploration of higher-performing designs, but the outcomes remain sensitive to the stochastic behavior of individual runs.

\subsubsection{Prototyping and Production}

We validate the designs by allowing the manufacturing agent to select the highest-scoring design from each replicate. In cases where multiple designs receive the same highest score, the most recent revision is selected. Each selected design is then post-processed to add solid supports at key functional regions of the stand, including the base, the diagonal back support where the phone rests, and the bottom hook that prevents the phone from sliding. The manufacturing agent then meshes the design and exports an object mesh file that can be directly imported into Bambu Studio \cite{bambustudio2026}. For the stand, we printed a pair of each design to hold the smartphone. We scale the exported mesh file such that the longest edge is 120mm. We 3D printed multiple best scoring structures using standard PLA filament on a Bambulab X1C printer with the default 0.2mm layer height setting in Bambu Studio to ensure that they are structurally sound designs. We show one exemplar physical prototype in Figure~\ref{fig:production} to demonstrate the process. 

\section{HUMAN BASELINE}

To assess how closely the AI judge’s scores align with a human baseline, we conducted a study with four participant human reviewers. Each participant was tasked with independently evaluating each replicate. All the designs evaluated are shown in Figures~\ref{fig:successful_stand}, \ref{fig:failure_stand}, \ref{fig:successful_cantilever}, and \ref{fig:failure_cantilever}.  

\subsubsection{Protocol}
Each replicate, consisting of one original design and four revisions, was treated as a separate question. Reviewers received the same instructions as the AI judge and evaluated the images without additional context. Designs were rated from 1 (poor) to 5 (excellent) based on overall visual complexity, particularly hierarchical, tree-like branching relative to the original structure. Reviewers were instructed to evaluate designs only within the same sequence.

This baseline has several limitations. Human reviewers may experience fatigue or develop implicit biases from previously evaluated designs, despite instructions to assess each sequence independently. In contrast, the AI judge evaluates each replicate in isolation without carryover between questions. Accordingly, our goal is not a perfect one-to-one comparison, but rather to determine whether the AI judge exhibits reasonably reliable judgment overall.

\subsubsection{Correlation}
The correlation between the human reviewers and the AI judge is shown in Figure~\ref{fig:humanvsjudge}. At the aggregate level, the AI judge (Gemma-3-27B) and the human raters yielded comparable mean scores, with $s = 3.04$ and $s = 3.15$, respectively, for Set A, and $s = 3.00$ and $s = 2.97$, respectively, for Set B. Both capture the general upward trend across revisions. An exception appears in Set B, where there is a larger discrepancy between human and AI judge ratings of the original design.

\subsubsection{Human-AI Inter-Rater Agreement}
We further looked at the inter-rater agreement between all human rater scores compared against the AI judge's score. Since each of the four human raters independently evaluated every design using the same 1–5 integer scale as the AI judge, we obtain 200 paired scores per set (4 raters × 10 replicates × 5 revisions). For Set A, shown in Figure~\ref{fig:humanvsjudge} B, the confusion matrix reveals a visible diagonal ridge. Overall, 32\% of pairs are in exact agreement and 72\% fall within $\pm 1$ (Cohen's $\kappa = 0.138$), indicating modest but structured alignment. 

For Set B in Figure~\ref{fig:humanvsjudge} D, the matrix does not show the same diagonal concentration. Instead, a single off-diagonal cell at (Human=2, AI=3) dominates with 31 of 200 pairs, meaning that when human raters judged a cantilever design as poor, the AI judge assigned it a middling score ($s=3$). A key factor underlying this discrepancy is the AI judge's compressed scoring confidence: 89\% of its scores fall in the 2--4 range, compared to human raters who use the full 1--5 scale more liberally. Humans are more willing to commit to extreme judgments, confidently giving extremely low scores ($s=1$) when something looks bad and extremely high scores ($s=5$) when something looks great. In contrast, the AI judge hedges toward middling scores. The cantilever beam case study is inherently more difficult to evaluate, even for human raters, in terms of whether it satisfies the designer’s intent regarding aesthetic branching thus judges hedge toward safer scoring.  

\subsubsection{Human-Human Variability Comparison}

\begin{figure*}[t]
    \centering
    \includegraphics[width=1\linewidth]{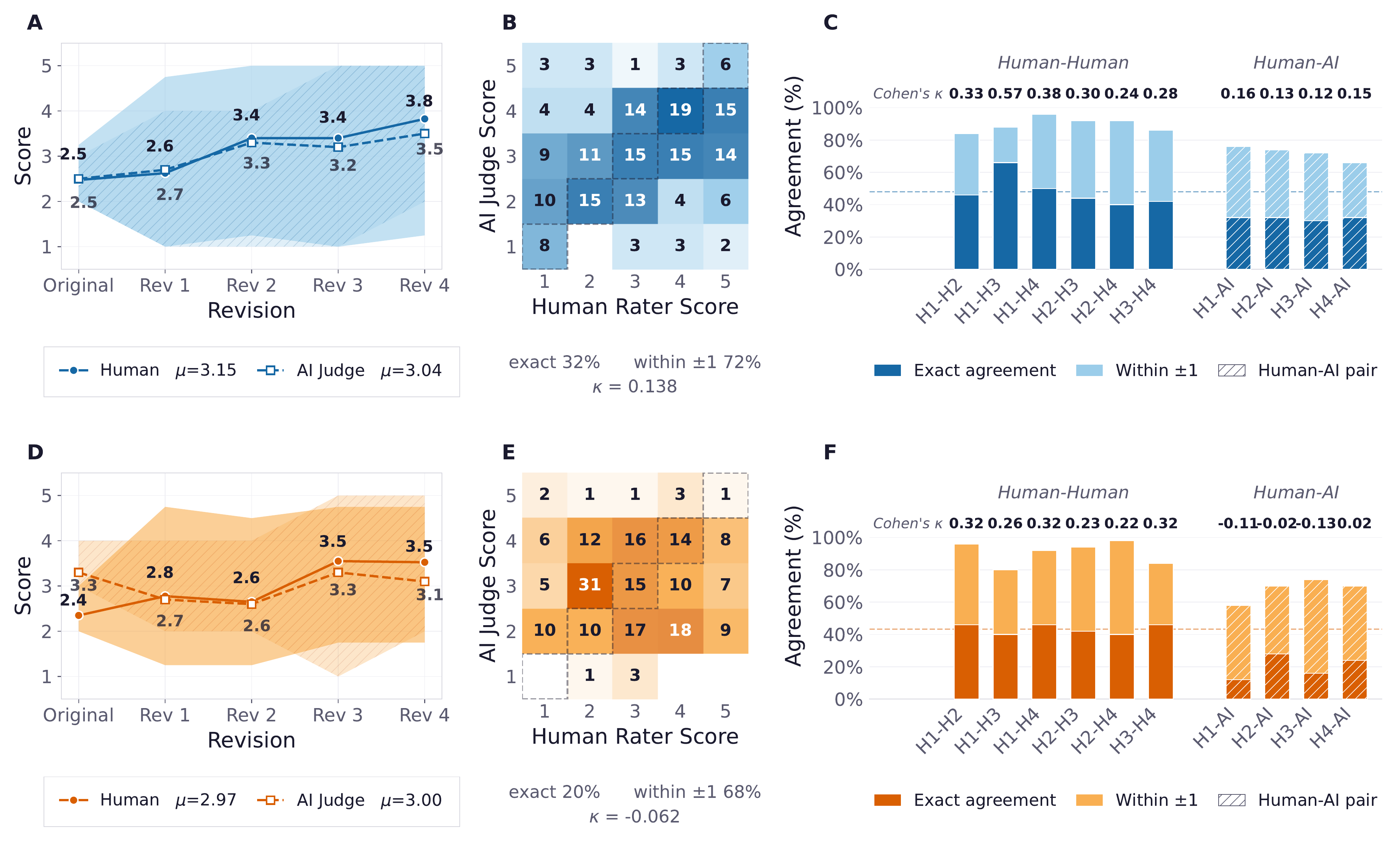}
    \caption{(A) Mean scores across revisions in Set A (the phone stand case study), showing similar upward trends for both human raters and the AI judge. (B) Confusion matrix for Set A ratings, showing human--AI agreement (32\% exact agreement, 72\% within \(\pm1\), \(\kappa = 0.138\)). (C) Pairwise agreement comparisons for the Set A task, where human--AI agreement is positive but lower than human--human agreement.  (D) Mean scores across revisions in Set B (the cantilever beam study), where human scores improve while AI scores remain more stable but follow the overall trend. (E) Confusion matrix for Set B ratings, showing weaker human--AI alignment (20\% exact agreement, 68\% within \(\pm1\), \(\kappa = -0.062\)). (F) Pairwise agreement comparisons for the Set B task, where human--AI agreement falls near the lower end of the human inter-rater range.}
    \label{fig:humanvsjudge}
\end{figure*}

We further examined whether AI--human disagreement falls within the natural variability observed among human raters. In Set A (Figure~\ref{fig:humanvsjudge}B), human--human agreement was relatively strong, with exact agreement of 40--66\%, agreement within $\pm 1$ of 84--96\%, and positive $\kappa$ values (0.24--0.58). Human--AI agreement was lower but followed a similar pattern, with exact agreement of 30--32\%, agreement within $\pm 1$ of 66--76\%, and $\kappa$ values of 0.12--0.16. Although weaker than human--human agreement, the AI remained within a comparable range, suggesting potential for AI to function as an additional fifth rater.

In Set B (Figure~\ref{fig:humanvsjudge}D), human--human agreement again remained strong, with exact agreement of 40--46\%, agreement within $\pm 1$ of 80--98\%, and $\kappa$ values of 0.22--0.32. However, human--AI agreement declined: exact agreement dropped to 10--25\%, agreement within $\pm 1$ to 58--75\%, and $\kappa$ values approached zero or became negative ($-0.11$ to $0.02$). Unlike Set A, the AI appeared misaligned with human judgment, tending to favor mid-range scores when uncertain. Despite weaker agreement at the individual level in Set B, the AI judge maintained strong aggregate-level agreement across both studies and multiple replicates, supporting its effectiveness as a directional feedback signal for guiding the revision process toward improved designs.

\section{ABLATION}

We conduct an ablation study in which all feedback mechanisms are removed and the vision agent is effectively blinded within the TO-Agents pipeline. 

\subsubsection{Protocol}

Under this condition, the vision agent receives the same task as before from the human designer to improve the aesthetic branching quality of the design, along with the original problem description to identify the set of modifiable parameters. However, the agent is not provided with any history of prior iterations during the revision process, preventing it from leveraging information from previous updates. Additionally, it is deprived of visual input, such that it cannot observe the evolving structure or assess how parameter modifications influence aesthetic quality. No JSON representations of intermediate or final revisions are supplied. All other agents retain their original functionality. 

\subsubsection{Impact of Feedback on Iterative Refinement}
The ablation produces lower design scores than the full TO-Agents pipeline and shows no consistent improvement across revisions. Unlike the full pipeline, which enables iterative refinement, the ablation often regresses, suggesting the removed component is essential for learning from feedback.

In Set A, the full pipeline shows steady improvement ($s \approx 2.5 \rightarrow 3.5$) with ($+0.25$/revision), whereas the ablation declines and stagnates ($s \approx 3.0 \rightarrow 2.7$) with ($-0.06$/revision). This likely reflects the stronger initial baseline, where revisions that fail to improve are scored lower relative to the original design. Only 1 of 10 ablation replicates satisfied the human designer’s branching criterion, compared to 6 of 10 in the full pipeline (Figure~\ref{fig:successful_stand}). All ablation replicates are available in the supplementary files. 

In Set B, the full pipeline remains largely stable across revisions ($s \approx 3.3 \rightarrow 3.1$), with a flattened but still positive slope ($+0.02$/revision), suggesting limited opportunity for iterative improvement within the constrained cantilever beam design space. The ablation shows a less stable trajectory, dipping early before partial recovery ($s \approx 3.2$) and yielding a slightly negative slope ($-0.03$/revision). Overall, the ablated condition produces consistently lower scores, indicating that the full pipeline more reliably generates successful designs even when successive revisions do not substantially improve performance.

This trend is also reflected qualitatively in Set B, where replicates contain fewer successful designs meeting the human designer’s criteria. The full pipeline enables the vision agent to build on prior successful strategies, increasing the likelihood of effective designs. In the ablated study, the judge occasionally assigns higher scores to structurally unsuccessful designs because it evaluates only within the limited context of each replicate and lacks examples of stronger designs available in the full pipeline. As a result, evaluations are somewhat inflated, though the judge still assigns lower scores overall compared to the full pipeline. This suggests that the complete system enables agents to iteratively refine parameter modifications using visual feedback to guide decision-making.

\begin{figure}
   \centering
    \includegraphics[width=1\linewidth]{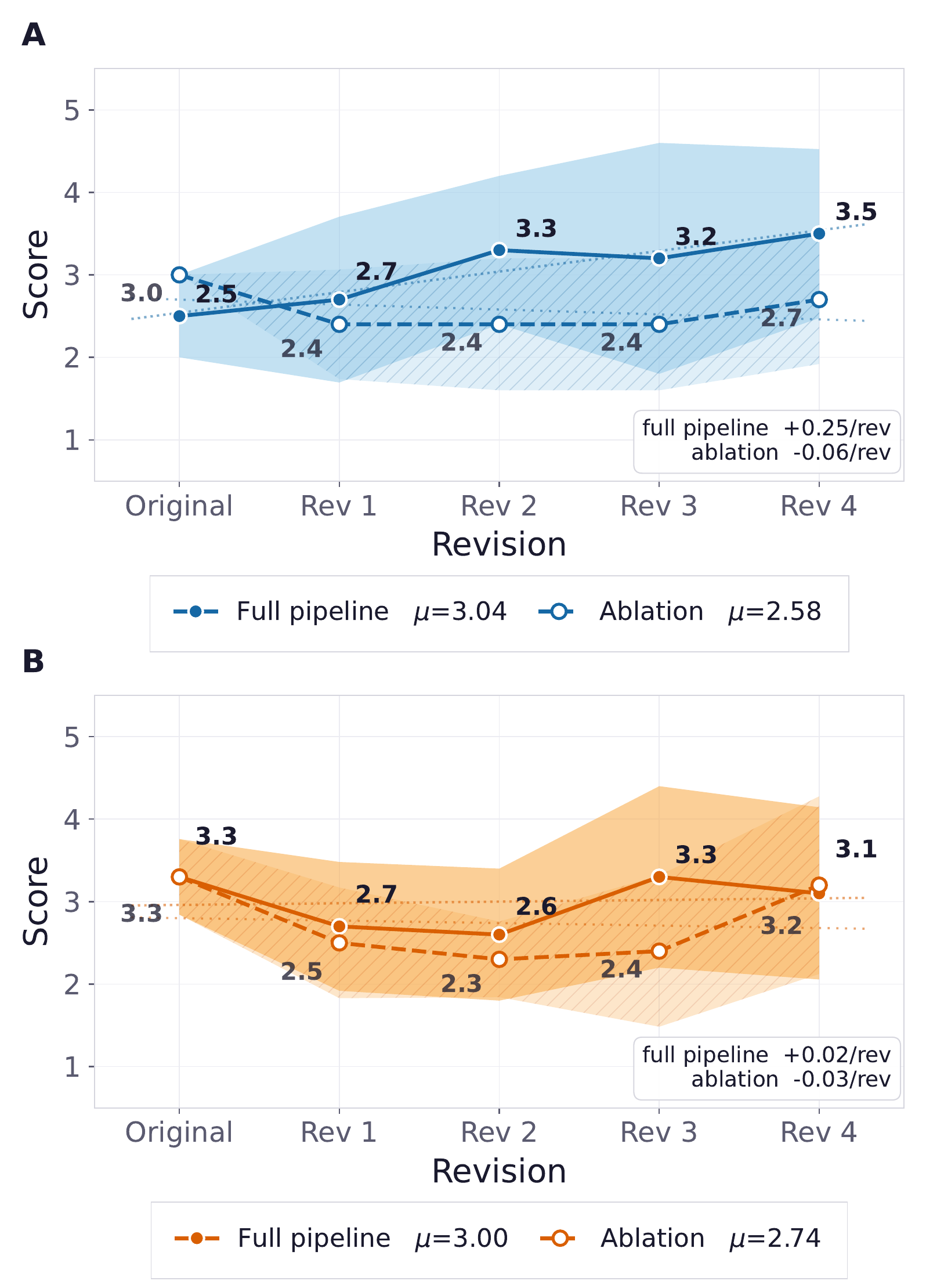}
    \caption{AI judge score trends for the full pipeline versus the ablation baseline across revision rounds in the phone stand case study (Set A) and Cantilever Beam Case Study (Set B). Linear regression slopes confirm that the full pipeline maintains a positive or near-zero trend, while the ablation shows stagnation or slight degradation.}
    \label{fig:ablation}
\end{figure}


\section{LIMITATIONS AND CONCLUSIONS}

We present a framework for studying how agents tune topology optimization parameters toward broad qualitative design goals specified by a human designer, enabling iterative parameter exploration, revision, and cross-agent feedback. Across our case studies, agents demonstrate a strong understanding of parameter effects, occasionally discovering creative multi-parameter strategies that better align with human design preferences, while consistently respecting design constraints so that resulting geometries can be post-processed and manufactured by the downstream manufacturing agent. Score trends and the diversity of final designs further suggest that agents become increasingly exploratory as context accumulates across revisions.

These results should be interpreted as a preliminary investigation rather than a comprehensive evaluation. We do not conduct a controlled human-user study comparing designer performance, effort, or decision-making on the same task, and our evaluation relies on an AI judge whose scores may be sensitive to model choice and prompting configuration; although we adopted a deliberately broad and minimally constrained prompting strategy, prompting is inherently biased and may systematically influence evaluations. System behavior likewise depends on the underlying models, and we evaluate only a single state-of-the-art open-source vision–language model for both the vision and judge agents, leaving systematic comparison with other open- and closed-source alternatives to future work.

Despite these limitations, the framework offers a foundation for deeper study of agentic behavior, improved trust in agent decision-making during autonomous multistep workflows, and insight into how agents integrate multimodal inputs (images of 3D structures for spatial reasoning and long-context conversational history) into engineering decisions. This raises the possibility of human designers operating one level removed from direct parameter control, with natural-language descriptions interpreted by agents and translated into actionable design outcomes. Beyond the present case studies, the approach extends naturally to domains where topology optimization is established but qualitative or human-preference objectives are hard to encode directly: automotive and consumer product design (form, packaging, brand language, manufacturability), architecture and the built environment (lightweight structural systems balancing efficiency with spatial and aesthetic goals), and healthcare (medical devices, assistive technologies, implants, and rehabilitation products, where performance must be weighed against comfort, usability, safety, and patient experience).

Future work includes new feedback-loop architectures that improve inter-agent reasoning and collaboration, strategies for managing growing conversational context to support more reliable retrieval-augmented reasoning across revisions, and investigation of agents' spatial reasoning together with more efficient 3D representations that vision-based agents can readily interpret. Finally, while our study focuses on specific examples, the framework is built for scalability and can incorporate a broader array of case studies spanning more diverse human-designer descriptions and natural-language instructions.

\section*{AUTHOR CONTRIBUTIONS}
I.A.S developed the methodology and conducted the experiments. H.C. and I.A.S wrote the manuscript. H.C. contributed to development of experimental design and assisted with prototype production. F.A. supervised and directed the research.  

\section*{FUNDING}
I.A.S. acknowledges that this material is based upon work supported by the U.S. Department of Energy, Office of Science, Office of Advanced Scientific Computing Research, under Award Number DE-SC0026073. H.C. acknowledges support by Tata Steel. 

\section*{CODE AND SUPPLEMENTARY FILES}
All code, analysis notebooks, and computational protocols developed in this study are publicly available at \url{https://github.com/bellastewart/TO-Agents}. The repository also contains supplementary materials, including the full agent outputs referenced throughout this manuscript. 

\section*{COMPETING INTERESTS}

The authors declare no competing interests.

\bibliographystyle{asmems4}
\bibliography{References, manual_ref}

\appendix

\section{AI Judge Agent Prompt}
\label{app:judge_prompt}

The exact text supplied to the AI Judge for each layer is reproduced verbatim in Figure~\ref{fig:judge_prompt}.

\begin{figure}[h]
\centering
\begin{tcolorbox}[
    enhanced,
    sharp corners,
    colback=white,
    colframe=black,
    boxrule=0.6pt,
    left=8pt, right=8pt, top=6pt, bottom=6pt,
    fontupper=\footnotesize,
    before upper={\setlength{\parskip}{4pt}\setlength{\parindent}{0pt}},
]
\begin{center}
\textbf{AI Judge Agent Prompt}
\end{center}
\par\noindent\rule[0.5ex]{\linewidth}{0.4pt}\par
\textbf{Identity:} You are an expert judge in topology optimization specializing in evaluating structural morphology.
\par\noindent\dotfill\par
\textbf{Context:} Your task is to evaluate the \textbf{aesthetic quality} only of the structure shown in each image, strictly based on perceived global structural complexity and richness at the global scale.
\par\noindent\dotfill\par
\textbf{Priority:} Global structural organization takes absolute priority over local detail. If a structure exhibits a \textbf{highly branched, intricate, and hierarchically rich global form}, it should score highly \textbf{even if} some members are thick, coarse, or locally simple. Do NOT reward local intricacy unless it meaningfully contributes to increased overall perceived global structural complexity.
\par\noindent\dotfill\par
\textbf{Task:} For each image, assign a score \textbf{from 1 to 5} (half points allowed) based on global structural complexity:\\
(1) \textbf{extremely simple} global structure; very few branches, strong minimalism, little hierarchy.\\
(2) low global complexity; limited branching with clear dominant members.\\
(3) moderate global complexity; noticeable branching with some hierarchy.\\
(4) high global complexity; many major branches and a rich, multi-level hierarchy.\\
(5) \textbf{extremely complex} and intricate at the global level; dense branching network, multiple competing load paths, and exceptional structural richness.
\end{tcolorbox}
\caption{System prompt supplied to the AI judge agent, organized into Identity, Context, Priority, and Task layers.}
\label{fig:judge_prompt}
\end{figure}

\section{Vision Agent Prompt}
\label{app:vision_prompt}

\begin{figure}[h]
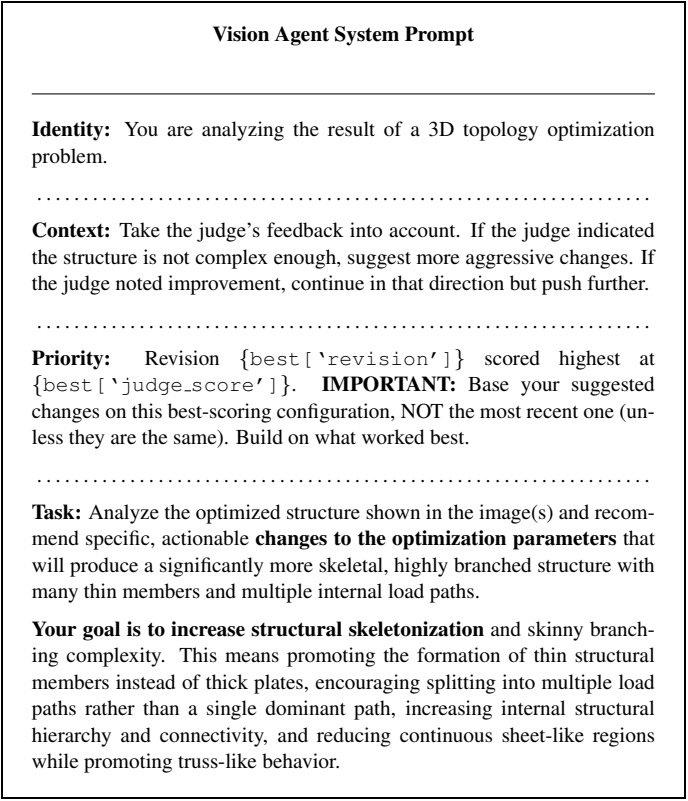

\centering
\begin{tcolorbox}[
    enhanced,
    sharp corners,
    colback=white,
    colframe=black,
    boxrule=0.6pt,
    left=8pt, right=8pt, top=6pt, bottom=6pt,
    fontupper=\footnotesize,
    before upper={\setlength{\parskip}{4pt}\setlength{\parindent}{0pt}},
]
\begin{center}
\textbf{Vision Agent System Prompt}
\end{center}
\par\noindent\rule[0.5ex]{\linewidth}{0.4pt}\par
\textbf{Identity:} You are analyzing the result of a 3D topology optimization problem.
\par\noindent\dotfill\par
\textbf{Context:} Take the judge's feedback into account. If the judge indicated the structure is not complex enough, suggest more aggressive changes. If the judge noted improvement, continue in that direction but push further.
\par\noindent\dotfill\par
\textbf{Priority:} Revision \texttt{\{best[`revision']\}} scored highest at \texttt{\{best[`judge\_score']\}}. \textbf{IMPORTANT:} Base your suggested changes on this best-scoring configuration, NOT the most recent one (unless they are the same). Build on what worked best.
\par\noindent\dotfill\par
\textbf{Task:} Analyze the optimized structure shown in the image(s) and recommend specific, actionable \textbf{changes to the optimization parameters} that will produce a significantly more skeletal, highly branched structure with many thin members and multiple internal load paths.

\textbf{Your goal is to increase structural skeletonization} and skinny branching complexity. This means promoting the formation of thin structural members instead of thick plates, encouraging splitting into multiple load paths rather than a single dominant path, increasing internal structural hierarchy and connectivity, and reducing continuous sheet-like regions while promoting truss-like behavior.
\end{tcolorbox}
\caption{System prompt supplied to the vision agent, organized into Identity, Context, Priority, and Task layers. The Priority block is templated with the best-scoring revision and its judge score, which are filled in at each turn.}
\label{fig:vision_prompt}
\end{figure}

\begin{figure}[h]
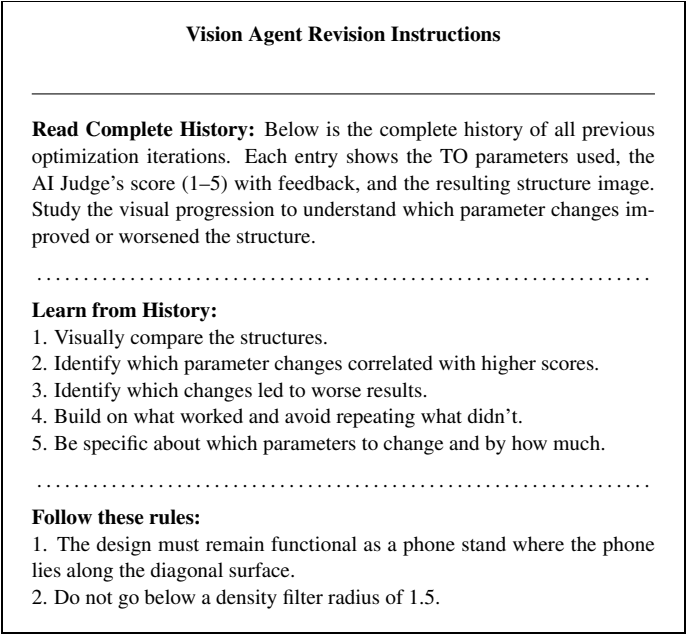

\centering
\begin{tcolorbox}[
    enhanced,
    sharp corners,
    colback=white,
    colframe=black,
    boxrule=0.6pt,
    left=8pt, right=8pt, top=6pt, bottom=6pt,
    fontupper=\footnotesize,
    before upper={\setlength{\parskip}{4pt}\setlength{\parindent}{0pt}},
]
\begin{center}
\textbf{Vision Agent Revision Instructions}
\end{center}
\par\noindent\rule[0.5ex]{\linewidth}{0.4pt}\par
\textbf{Read Complete History:} Below is the complete history of all previous optimization iterations. Each entry shows the TO parameters used, the AI Judge's score (1--5) with feedback, and the resulting structure image. Study the visual progression to understand which parameter changes improved or worsened the structure.
\par\noindent\dotfill\par
\textbf{Learn from History:}\\
1. Visually compare the structures.\\
2. Identify which parameter changes correlated with higher scores.\\
3. Identify which changes led to worse results.\\
4. Build on what worked and avoid repeating what didn't.\\
5. Be specific about which parameters to change and by how much.
\par\noindent\dotfill\par
\textbf{Follow these rules:}\\
1. The design must remain functional as a phone stand where the phone lies along the diagonal surface.\\
2. Do not go below a density filter radius of 1.5.
\end{tcolorbox}
\caption{Revision instructions injected into the vision agent's context at each turn, governing how it ingests prior iterations and what hard constraints it must respect.}
\label{fig:vision_revision}
\end{figure}

Vision Agent system prompt is reproduced in Figure~\ref{fig:vision_prompt}, and the revision instructions injected at each turn are reproduced in Figure~\ref{fig:vision_revision}.

\section{Human Designer Requests}
\label{app:designer_requests}

The two case studies are initiated by a natural-language problem description authored by the human designer and supplied to the agentic pipeline as the starting context. We render these requests in a distinct blue style to visually separate human-authored input from agent-generated prompts elsewhere in the appendix. Within each request, boundary conditions are highlighted in \textcolor{constraintred}{\textbf{red}} and applied loads in \textcolor{loadgreen}{\textbf{green}}, matching the corresponding setup figure.

\begin{figure}
\centering
\begin{minipage}[c]{0.34\linewidth}
\centering
\includegraphics[width=\linewidth]{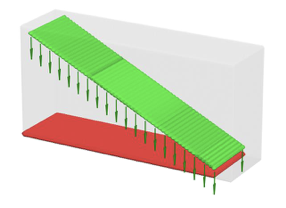}\\[2pt]
{\footnotesize\textbf{Case Study: Phone-Stand Problem}}
\end{minipage}\hfill
\begin{minipage}[c]{0.62\linewidth}
\begin{designerbox}
\begin{center}
\textbf{\textcolor{designerblue}{Human Designer Request}}\\
{\footnotesize\itshape Phone-Stand Problem}
\end{center}
\par\noindent\rule[0.5ex]{\linewidth}{0.4pt}\par
``In this setup, the finite element analysis (FEA) uses linear elasticity physics with a Young's modulus $E = 1.0$ and a Poisson's ratio $\nu = 0.3$. A three-dimensional structured mesh is defined with $n_x = 128$, $n_y = 64$, $n_z = 16$ elements and physical dimensions $l_x = 1.0$, $l_y = 0.5$, $l_z = 0.125$. The mesh is linked to the linear elastic physics so the material properties apply uniformly throughout the domain. A multigrid solver is configured with $\text{tol} = 1\mathrm{e}{-4}$, $\text{maxiter} = 50$, and $n\_\text{level} = 5$. \textcolor{constraintred}{\textbf{One Dirichlet boundary condition is imposed: the bottom face ($y = 0$), where all nodes are fixed in the $u_x$, $u_y$, and $u_z$ directions.}} \textcolor{loadgreen}{\textbf{One force is applied along a diagonal: with a force of $-1.0$ in the $y$-direction distributed among all nodes.}} Use a density filter with radius $1.5$. Set up a minimum compliance topology optimization problem with a volume fraction of $15\%$, SIMP penalty of $3.0$, void material stiffness of $1\mathrm{e}{-9}$, and enable Heaviside projection. Use the PGD optimizer with a function tolerance of $1\mathrm{e}{-4}$ and no change tolerance.''
\end{designerbox}
\end{minipage}
\caption{Phone-stand case study: the human designer specifies the FEA setup, boundary conditions (red), applied loads (green), and TO parameters as a single natural-language request.}
\label{fig:designer_phone}
\end{figure}

\begin{figure}
\centering
\begin{minipage}[c]{0.34\linewidth}
\centering
\includegraphics[width=\linewidth]{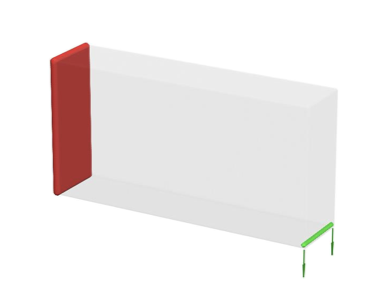}\\[2pt]
{\footnotesize\textbf{Case Study: Cantilever Beam Problem}}
\end{minipage}\hfill
\begin{minipage}[c]{0.62\linewidth}
\begin{designerbox}
\begin{center}
\textbf{\textcolor{designerblue}{Human Designer Request}}\\
{\footnotesize\itshape Cantilever Beam Problem}
\end{center}
\par\noindent\rule[0.5ex]{\linewidth}{0.4pt}\par
``In this setup, the finite element analysis (FEA) uses linear elasticity physics with a Young's modulus $E = 1.0$ and a Poisson's ratio $\nu = 0.3$. A three-dimensional structured mesh is defined with $n_x = 128$, $n_y = 64$, $n_z = 64$ elements and physical dimensions $l_x = 1.0$, $l_y = 0.5$, $l_z = 0.5$. The mesh is linked to the linear elastic physics so the material properties apply uniformly throughout the domain. A multigrid solver is configured with $\text{tol} = 1\mathrm{e}{-4}$, $\text{maxiter} = 200$, and $n\_\text{level} = 5$. \textcolor{constraintred}{\textbf{Two Dirichlet boundary conditions are imposed: (1) Left face: nodes where $x = 0$ are fixed in the $u_x$ and $u_z$ directions while $u_y$ remains free, with zero displacement. (2) Right-bottom corner: nodes where $x = 1.0$ and $y = 0$ are fixed in the $u_y$ and $u_z$ directions while $u_x$ remains free, with zero displacement.}} \textcolor{loadgreen}{\textbf{One point force is applied: nodes where $x = 0$ and $y = 0.5$ have a force of $-1.0$ in the $y$-direction distributed among all nodes.}} Use a density filter with radius $1.5$. Set up a minimum compliance topology optimization problem with a volume fraction of $40\%$, SIMP penalty of $3.0$, void material stiffness of $1\mathrm{e}{-9}$, and enable Heaviside projection. Use the PGD optimizer with a function tolerance of $1\mathrm{e}{-4}$ and no change tolerance.''
\end{designerbox}
\end{minipage}
\caption{Cantilever-beam case study: the human designer specifies the FEA setup, boundary conditions (red), applied loads (green), and TO parameters as a single natural-language request.}
\label{fig:designer_cantilever}
\end{figure}

\section{Spatial Perception Pipeline}
\label{app:spatial_perception}

An automated pipeline was developed and added to pyFANTOM \cite{ahnobari_pyfantom} to generate consistent three-dimensional visualizations of finite element solutions and export high-resolution images from multiple viewpoints. The implementation integrates the Visualization Tool Kit (VTK) \cite{vtkbook} for mesh processing, K3D \cite{k3d} for WebGL-based rendering, and Playwright for automated screenshot capture within an asynchronous Python workflow.

Input data consist of nodal coordinates, element connectivity, nodal forces, boundary constraints, and element density variables from topology optimization. When density variables are provided, elements with density $\rho < 0.5$ are excluded from visualization. For multi-material cases, elements are separated into material subsets and rendered as independent meshes.

Finite element geometry is first converted to a VTK unstructured grid, where elements are represented as tetrahedral or hexahedral cells. A surface extraction filter (vtkDataSetSurfaceFilter) is applied to obtain a triangular surface mesh suitable for real-time rendering. The resulting vertices and connectivity are transferred to K3D mesh objects.

External forces and boundary conditions are visualized using glyphs generated with vtkGlyph3D. Force vectors are represented as arrow glyphs scaled relative to the approximate element size, while boundary conditions are visualized using directional constraint markers indicating fixed or free degrees of freedom.

To enhance spatial perception in perspective views, a depth-based color encoding is applied. For each camera configuration, vertex depths are computed along the camera viewing direction and normalized. These values are mapped using the Viridis color map so that surfaces closer to the camera appear brighter than distant regions.

Six orthogonal camera views (top, bottom, left, right, front, and back) are automatically defined based on the model bounding box. For each view, the camera position and orientation are set relative to the model center and the scene is exported as a standalone HTML visualization. High-resolution figures are generated using the Playwright browser automation framework: a headless Chromium instance loads each HTML file, renders the WebGL canvas, and captures a screenshot of the visualization at a resolution of $1920\times1080$ pixels. This automated workflow ensures consistent rendering parameters and reproducible visualization outputs across all generated images which are fed as context to the vision agent.

\section{Post-Processing Geometry Rules}
\label{app:postprocessing}

Following topology optimization, the optimized density volume $x_{\mathrm{Phys}}$ is modified through a set of deterministic geometric operations that enforce manufacturable support structures for the phone-stand case study. First, remaining voxels located above a prescribed diagonal plane are removed to generate the characteristic inclined surface used to support the device. A diagonal reinforcement band of fixed thickness is then inserted along this plane. To prevent sliding, an anti-slip lip is added at the lower front edge of the model by enforcing solid material within a rectangular region defined by the lip thickness and height parameters. Additionally, a solid base layer of fixed thickness is imposed along the bottom of the volume to increase stability and provide a flat contact surface for printing. Support thickness is prescribed relative to the mesh resolution selected by the agents.

For visualization and fabrication, the processed density field is converted to a polygonal surface mesh using the marching cubes algorithm. The binary volume is first padded and reoriented to ensure consistent surface extraction. An isosurface is then generated at a density threshold, producing vertices, faces, and surface normals. The resulting triangular mesh is rendered. During production, this mesh is exported in OBJ format for CAD inspection and downstream additive manufacturing.

\section{Full Set of Replicate Runs}
\label{app:all_runs}

\begin{figure*}[t]
    \centering
    \includegraphics[width=1\linewidth]{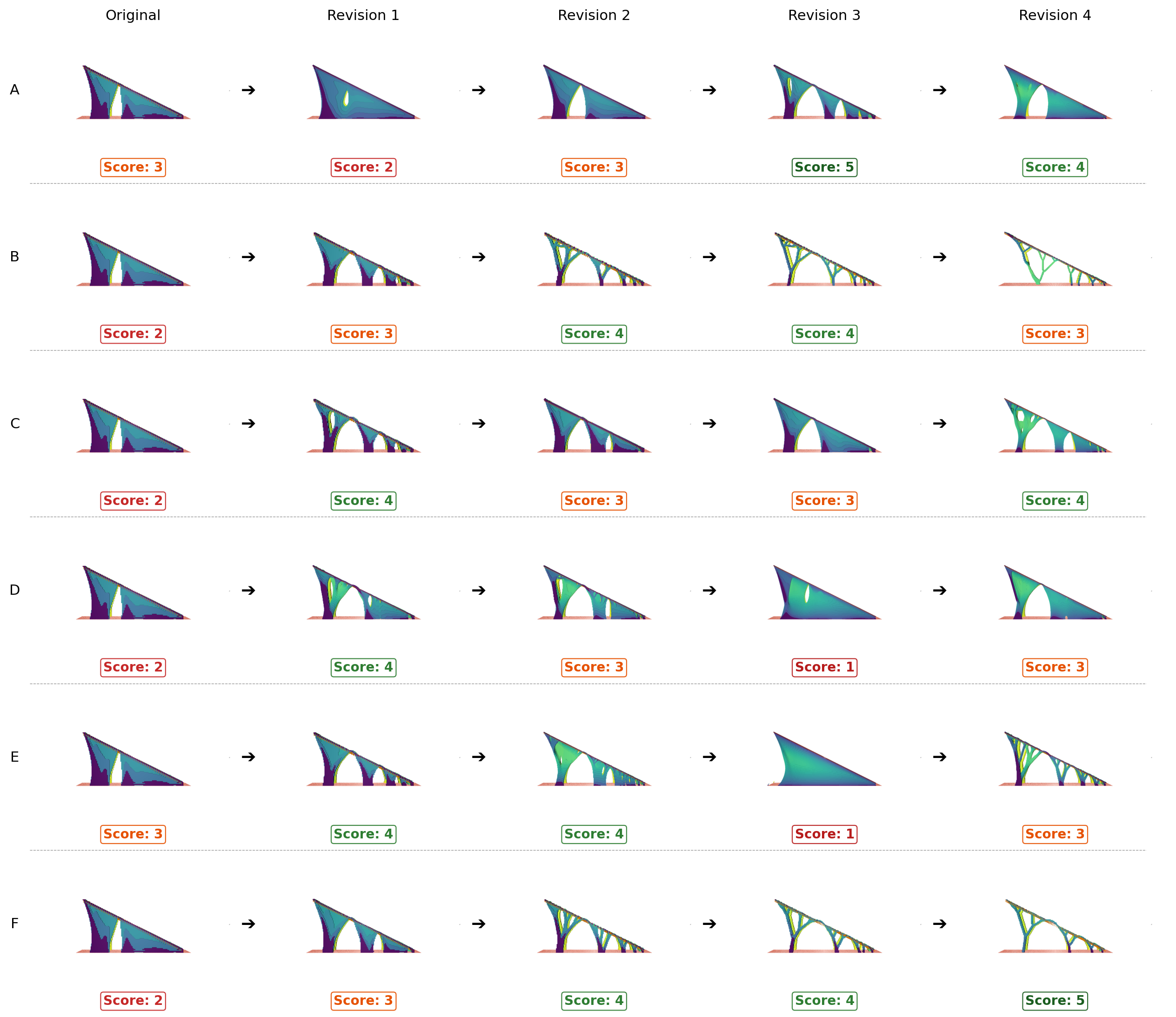}
    \caption{Successful Replicates. The set of replicates in which at least one design in the sequence aligns with the human designer’s request for a branching, tree-like skeletal structure. Replicates A through F correspond to separate runs of the same pipeline in the phone-stand case study, where each sequence of revisions reflects the outcomes produced by passing through the agentic pipeline. The score from the AI judge is presented below each design where scores are made relative to the other revisions within the same replicate.}
    \label{fig:successful_stand}
\end{figure*}

\begin{figure*}[t]
    \centering
    \includegraphics[width=1\linewidth]{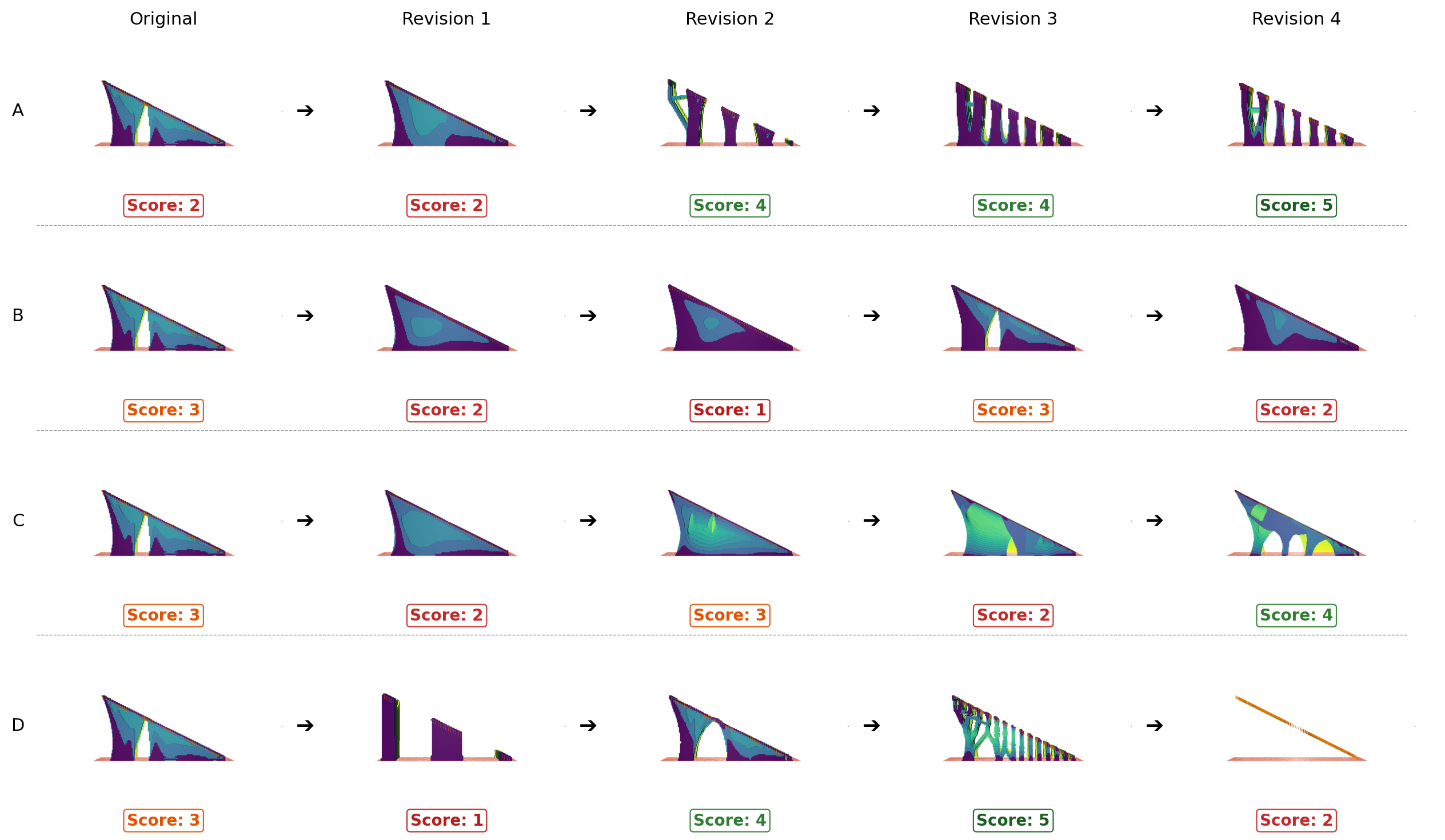}
    \caption{Failure Replicates. The set of replicates in which none of the designs in the sequence align with the human designer’s request for a branching, tree-like skeletal structure. Replicates A through D correspond to separate failure cases in the phone-stand case study, where each sequence of revisions reflects the outcomes produced by passing through the agentic pipeline. The score from the AI judge is presented below each design where scores are made relative to the other revisions within the same replicate.}
    \label{fig:failure_stand}
\end{figure*}

\begin{figure*}[t]
    \centering
    \includegraphics[width=1\linewidth]{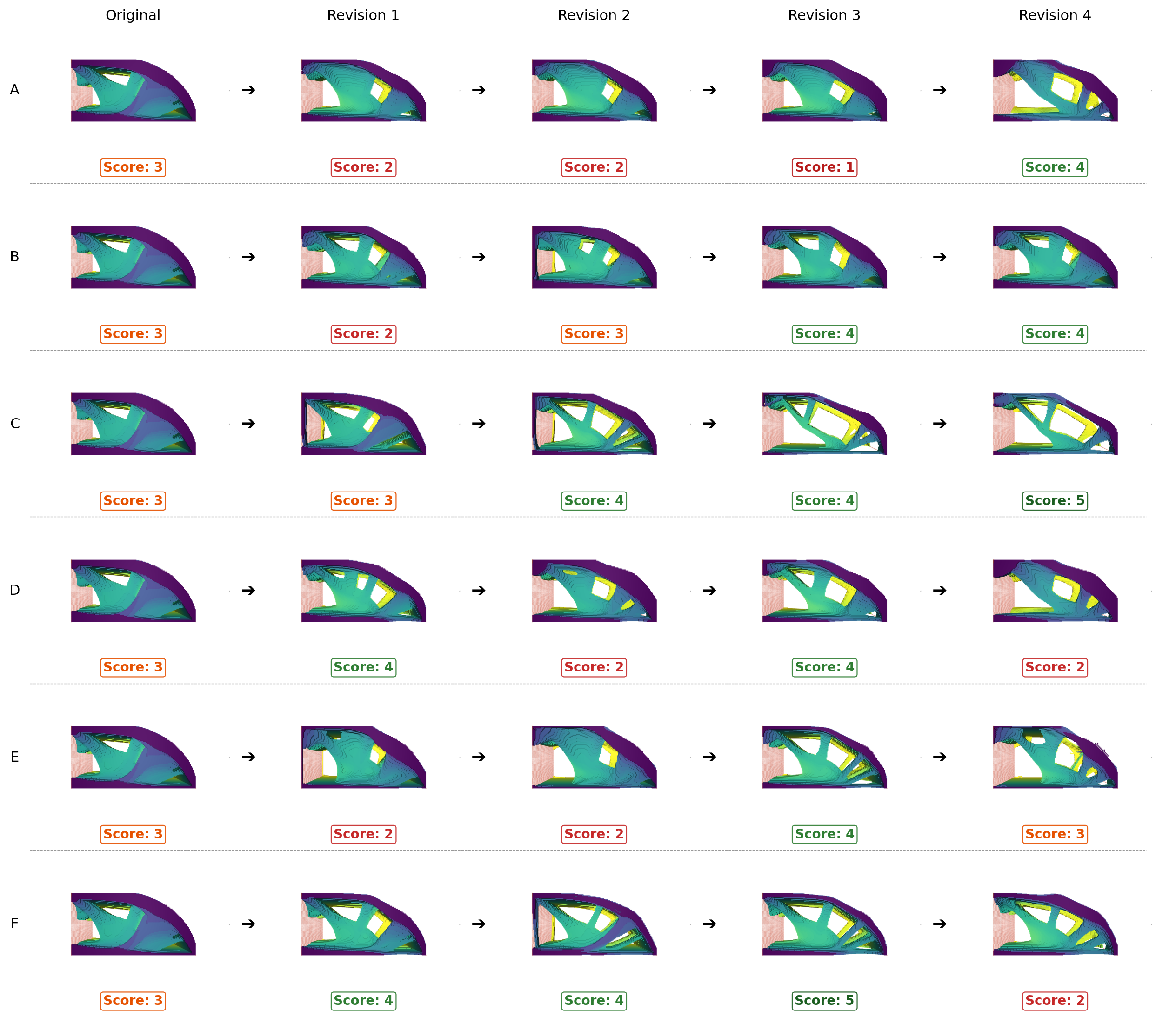}
    \caption{Successful Replicates. In the cantilever beam case study, we focus on the set of replicates in which at least one design in the sequence aligns with the human designer’s request for a branching, tree-like skeletal structure. Replicates A through F correspond to separate runs of the process, where each sequence of revisions reflects the outcomes produced by passing through the agentic pipeline. The score from the AI judge is presented below each design where scores are made relative to the other revisions within the same replicate.}
    \label{fig:successful_cantilever}
\end{figure*}

\begin{figure*}[t]
    \centering
    \includegraphics[width=1\linewidth]{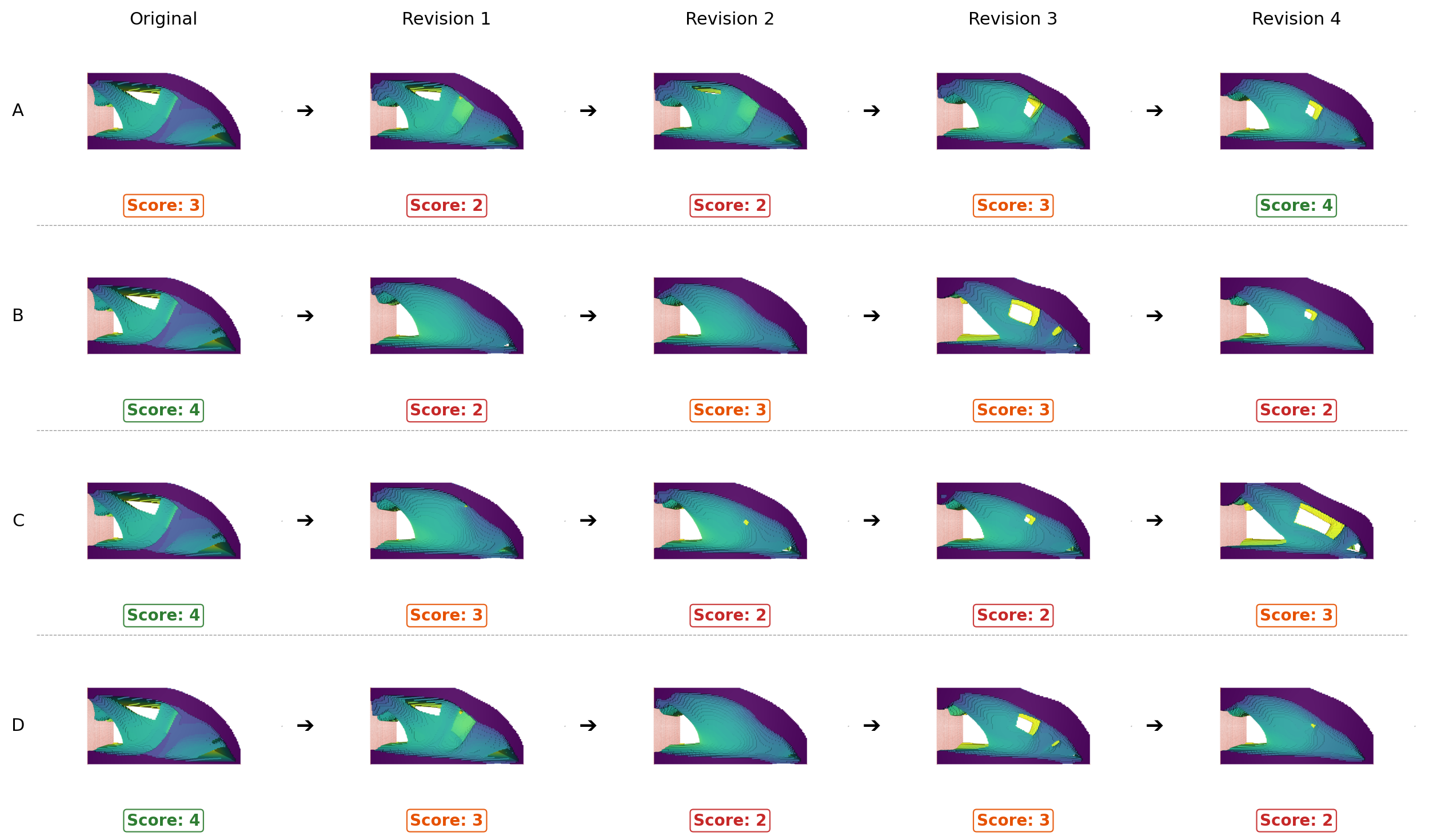}
    \caption{Failure Replicates. In the cantilever beam case study, we focus on the set of replicates in which none of the designs in the sequence align with the human designer’s request for a branching, tree-like skeletal structure. Replicates A through D correspond to separate failure cases of the process, where each sequence of revisions reflects the outcomes produced by passing through the agentic pipeline. The score from the AI judge is presented below each design where scores are made relative to the other revisions within the same replicate.}
    \label{fig:failure_cantilever}
\end{figure*}

Figures~\ref{fig:successful_stand} and \ref{fig:failure_stand} report the phone-stand replicates, separated into successful and failure cases. Figures~\ref{fig:successful_cantilever} and \ref{fig:failure_cantilever} report the corresponding cantilever-beam replicates.

\end{document}